\documentclass[a4paper,fleqn]{cas-dc}

\usepackage[numbers]{natbib}

\def\tsc#1{\csdef{#1}{\textsc{\lowercase{#1}}\xspace}}
\tsc{WGM}
\tsc{QE}
\tsc{EP}
\tsc{PMS}
\tsc{BEC}
\tsc{DE}

\usepackage{pdflscape} 
\usepackage{rotating}
\usepackage{booktabs}  
\usepackage{siunitx}  
\usepackage{longtable}
\usepackage{adjustbox}
\usepackage{amssymb}
\usepackage{pifont}
\newcommand{\cmark}{\ding{51}}%
\newcommand{\xmark}{\ding{55}}%

\begin{document}
\let\WriteBookmarks\relax
\def\floatpagepagefraction{1}
\def\textpagefraction{.001}
\shorttitle{Transformer-based inpainting}
\shortauthors{Elharrouss et~al.}

\title [mode = title]{Transformer-based Image and Video Inpainting: Current Challenges and Future Directions }

\author[1]{Omar Elharrouss}[]
\cormark[1]
\ead{Omar Elharrouss}

\author[1]{Rafat Damseh}[]
\ead{rdamseh@uaeu.ac.ae}

\author[2]{Abdelkader Nasreddine Belkacem}
\ead{belkacem@uaeu.ac.ae}

\author[1]{Elarbi Badidi}
\ead{ebadidi@uaeu.ac.ae}

\author[2]{Abderrahmane Lakas}
\ead{alakas@uaeu.ac.ae}

\address[1]{Department of Computer Science and Software Engineering, College of Information Technology, United Arab Emirates University.}
\address[2]{Department of Computer and Network Engineering, College of Information Technology, United Arab Emirates University.}

\cortext[cor1]{Corresponding author}

\begin{abstract}
Image inpainting is currently a hot topic within the field of computer vision. It offers a viable solution for various applications, including photographic restoration, video editing, and medical imaging. Deep learning advancements, notably convolutional neural networks (CNNs) and generative adversarial networks (GANs), have significantly enhanced the inpainting task with an improved capability to fill missing or damaged regions in an image or video through the incorporation of contextually appropriate details. These advancements have improved other aspects, including efficiency, information preservation, and achieving both realistic textures and structures. Recently, visual transformers have been exploited and offer some improvements to image or video inpainting. The advent of transformer-based architectures, which were initially designed for natural language processing, has also been integrated into computer vision tasks. These methods utilize self-attention mechanisms that excel in capturing long-range dependencies within data; therefore, they are particularly effective for tasks requiring a comprehensive understanding of the global context of an image or video. In this paper, we provide a comprehensive review of the current image or video inpainting approaches, with a specific focus on transformer-based techniques, with the goal to highlight the significant improvements and provide a guideline for new researchers in the field of image or video inpainting using visual transformers. We categorized the transformer-based techniques by their architectural configurations, types of damage, and performance metrics. Furthermore, we present an organized synthesis of the current challenges, and suggest directions for future research in the field of image or video inpainting.
\end{abstract}





\maketitle

\section{Introduction}
Image inpainting is a fundamental task in computer vision and image processing that involves the restoration or completion of missing or damaged regions within an image. Image inpainting offers a significant solution for many tasks including photography restoration, video editing, and medical imaging. Over the years, different techniques have been developed to address image or video inpainting, which range from traditional methods based on patch-based or exemplar-based approaches to recent deep learning-based techniques \cite{i1}. Patch-based image inpainting functions by searching for patches (small regions) within the image to fill in the missing or damaged parts \cite{i2}. The algorithm looks for the best matching patches that align with the boundary conditions of the damaged area and uses these patches to reconstruct the missing parts. It is particularly effective for small, damaged regions. The exemplar-based technique is an extension of the patch-based approach that incorporates additional information, such as texture and structure, into the selection process \cite{i3}. These methods prioritize patches that fit with the surrounding region with similar structural or textural information, resulting in the inpainting results being more coherent with the overall image content.
Recently, the advent of convolutional neural networks (CNNs) has revolutionized the field of image inpainting by training inpainting models on large-scale datasets \cite{i4,i41,i42,i43}. The learning capabilities of neural networks are forced to effectively capture the contextual information to fill the missing parts in the image whilst conserving the coherence and quality of the image. These approaches have demonstrated remarkable performance improvements over traditional methods, particularly in handling complex and large-scale inpainting tasks. In addition to CNN-based models, advanced techniques, such as generative adversarial networks (GANs), have been employed to generate high-quality inpainted images and to maintain the integrity of the original image in terms of the structural and textural components. Additionally, these techniques are powerful in handling complex textures and large missing regions in an image.
Furthermore, the use of transformer-based architectures, initially proposed for natural language processing tasks, has gained significant interest in the field of computer vision including image inpainting \cite{i5}. Transformers with self- attention mechanisms succeed at capturing long-range dependencies, making them suitable for the tasks that require a global context understanding, such as image inpainting \cite{i6}. Transformer-based inpainting methods allow the capture of contextual information from the entire image for accurate and coherent inpainting results.
In this paper, we aim to provide a comprehensive overview
of transformer-based image or video inpainting methods by the category of approaches, architectures, and performance characteristics of these approaches.

We present a selection of influential and essential algorithms from prestigious journals and conferences. The focus of this study is on contemporary transformer-based image or video inpainting methodologies, which can provide a deeper understanding of the advancements in image or video inpainting. Additionally, a discussion of recent advancements, challenges, and future research directions in the field of image or video inpainting using transformer-based methodologies is provided. The content of this review is presented as follows:
\begin{itemize}
     \item Summarization of the existing surveys.
    \item A description of different types of damage.
    \item 	Classification of transformer-based image or video inpainting methods.
    \item 	Public datasets deployed to evaluate the image or video inpainting methods.
   \item  Evaluation metrics are described with various comparisons of the most significant works.
   \item Current challenges and future directions.
\end{itemize}
The remainder of the paper is organized as follows. An overview of the related studies, the scope, and previous surveys are conducted in Section 2. A taxonomy of the image and video inpainting is presented in Section 3. Video inpainting methods are discussed in Section 4. Used loss functions are presented in Section 5. Public datasets are briefly described in Section 6 before presenting the evaluation metrics and various comparisons of the most significant image or video inpainting methods in Section 7.  Current challenges are presented in Section 8. Finally, a conclusion is provided in Section 9.


\section{Related previous reviews and surveys}

In the literature, reviews and surveys for image or video inpainting with different techniques and for different purposes were searched \cite{r1,r2,r3,r4,r5,r6,r7,r8,r9,r10,r11,r12}. 

These reviews can be categorized based on the data used in inpainting, such as scratched pictures inpainting  \cite{r1}, Depth images inpainting using 3D to 2D representations \cite{r2,r3}, RGB images \cite{r4,r5}, or forensics imaging \cite{r8,r81}. In addition, the techniques-based reviews can be divided into two categories, including traditional-based methods and deep learning-based approaches. The first reviews of traditional techniques used in image or video inpainting, including texture and patch-based techniques, are proposed in  \cite{r1}, \cite{r2}, and \cite{r3}.In some papers, the authors briefly discuss the traditional methods and then the deep learning methods, as in \cite{r9}. For the deep learning-based reviews which were mostly published during the last 5 years, papers included those based on CNNs \cite{r4,r6,r10,r12}, while \cite{r4,r5,r9,r11} focused on GANs. A summarization of the proposed survey is provided in Table \ref{surveys}.

\begin{table*}
\begin{center}
\caption{Summarization of crowd counting methods. Journal paper type is abbreviated to J and conference paper is abbreviated to Conf.}
\label{surveys}
\begin{tabular}{|c|c|p{1.5cm}|p{1,5cm}|p{11cm}|}
\hline 
\textbf{Reference} &	\textbf{Year }  &\textbf{Subject}& \textbf{Tech}   &\textbf{Brief description} \\
\hline
\cite{r1} & 2015&  Old pictures & Patch and texture-based & The paper presents the techniques used neighboring pixels and texture to restore missing or corrupted regions convincingly on old pictures. The various methods are discussed with their pros and cons as well as a comparative review of these techniques.\\
\hline

 \cite{r2} & 2016 &  Depth map inpainting & Structure and Texture-based & 
 The paper discusses two main types of inpainting methods including 2D image inpainting and depth map inpainting. 2D inpainting restores corrupted parts using structural or textural methods, while depth map inpainting enhances 3D visualization effects. \\
\hline

\cite{r3} &  2018&   Depth image & Traditional & 
Depth Image Based Rendering (DIBR) uses a gray-level depth map to shift pixels in a 2D image. However, this can leave "empty" areas that degrade image quality. Inpainting techniques are then used to fill these gaps. The paper surveys different image inpainting methods to address dis-occlusions in Depth Image Rendering.\\
\hline
\cite{r4} &2020 &   RGB images & CNN, GAN &
The paper reviews existing approaches categorized into sequential-based, CNN-based, and GAN-based methods. Each category includes methods tailored for different types of image distortion. The paper also presents available datasets and evaluates the performance of these methods on various distortions. \\
\hline

 \cite{r5} & 2020 &  RGB images & GAN&
The paper discusses recent advancements in Generative Adversarial Networks (GANs) within image processing, editing, and inpainting. It analyzes methods used in these applications, discusses challenges faced by GANs, and suggests future research directions. Overall, it provides insights into GAN research and its diverse applications.\\
\hline

 \cite{r6} & 2021 &  RGB images & Traditional and Deep learning &
 The paper provides an overview of image inpainting methods over the past decade, highlighting the shift from traditional restoration to digital techniques. It categorizes traditional methods into five sub-categories and reviews deep learning methods such as Convolutional Neural Networks and Generative Adversarial Networks. \\
\hline

 \cite{r7} & 2021 &  Facial & Various &
The paper explores automatic facial wrinkles detection and inpainting algorithms, emphasizing wrinkles as a key sign of aging. It surveys computer vision techniques and reviews datasets, detection, and inpainting algorithms. \\
\hline


\cite{r8} & 2022 & Forensic & Various &
The article examines the advancements in image painting technology and its implications for image forgery. It discusses the development of image forensics to detect forged images, noting current limitations compared to inpainting technology. .\\
\hline

\cite{r9} & 2022 &  RGB images & CNN, GAN & 
The paper discusses traditional methods briefly and then delves into deep learning-based approaches, particularly those using generative adversarial network models. The paper covers the problems addressed, advantages and disadvantages, areas for improvement, and analyzes challenges in image painting.\\
\hline

\cite{r10} &2022 &  RGB  & CNN &  
This paper reviews image similarity measures used for inpainting and super-resolution. It categorizes existing methods, discusses their applications in inpainting with deep learning, and compares their pros and cons.\\
\hline

\cite{r11} &2023 &   Loss-function datasets & Deep learning  &

This survey explores the advancements in deep learning-based image inpainting based on network structures, and loss functions. It also summarizes available open-source codes, datasets, and metrics. Real-world applications are discussed, along with performance analysis of different algorithms.\\
\hline

\cite{r12}&2023 & RGB images & Fusion CNNs &
This article reviews deep learning-based image inpainting, highlighting its significance in computer vision. It examines progress over the past 15 years, neural network structures, fusion methods, and different inpainting tasks.\\
\hline

\end{tabular}
\end{center}
\end{table*}

With the introduction of transformer techniques in the field of computer vision, studies have taken preference for the proposed architectures, in addition to the improvements added, especially for image or video inpainting. Unlike previous surveys, our work is focused on the transformer-based techniques for image or video inpainting. We summarize the existing image or video inpainting models from different aspects and list some of the effective approaches in terms of qualitative results on severe image or video inpainting datasets.
Furthermore, we describe and analyze the most commonly proposed architectures in terms of the technique used and solved challenges in image inpainting. Finally, we list the open issues and challenges for image inpainting, in addition to the future directions. Through this survey, we expect to make reasonable inferences and predictions for the future development of image or video inpainting, and provide feasible solutions and guidance for the problem of image processing in other domains.

\section{Taxonomy of image or video inpainting}

In this section, we review transformer-based image inpainting algorithms in the following taxonomies. First, we discuss the different types of damages (masks) in the image or video. Then, we present the different types of transformer-based architectures for image or video inpainting in detail. The important models are described in chronological order.
\begin{figure*}[t!]
\centering
  \begin{tabular}[b]{c}
    \includegraphics[width=1\linewidth]{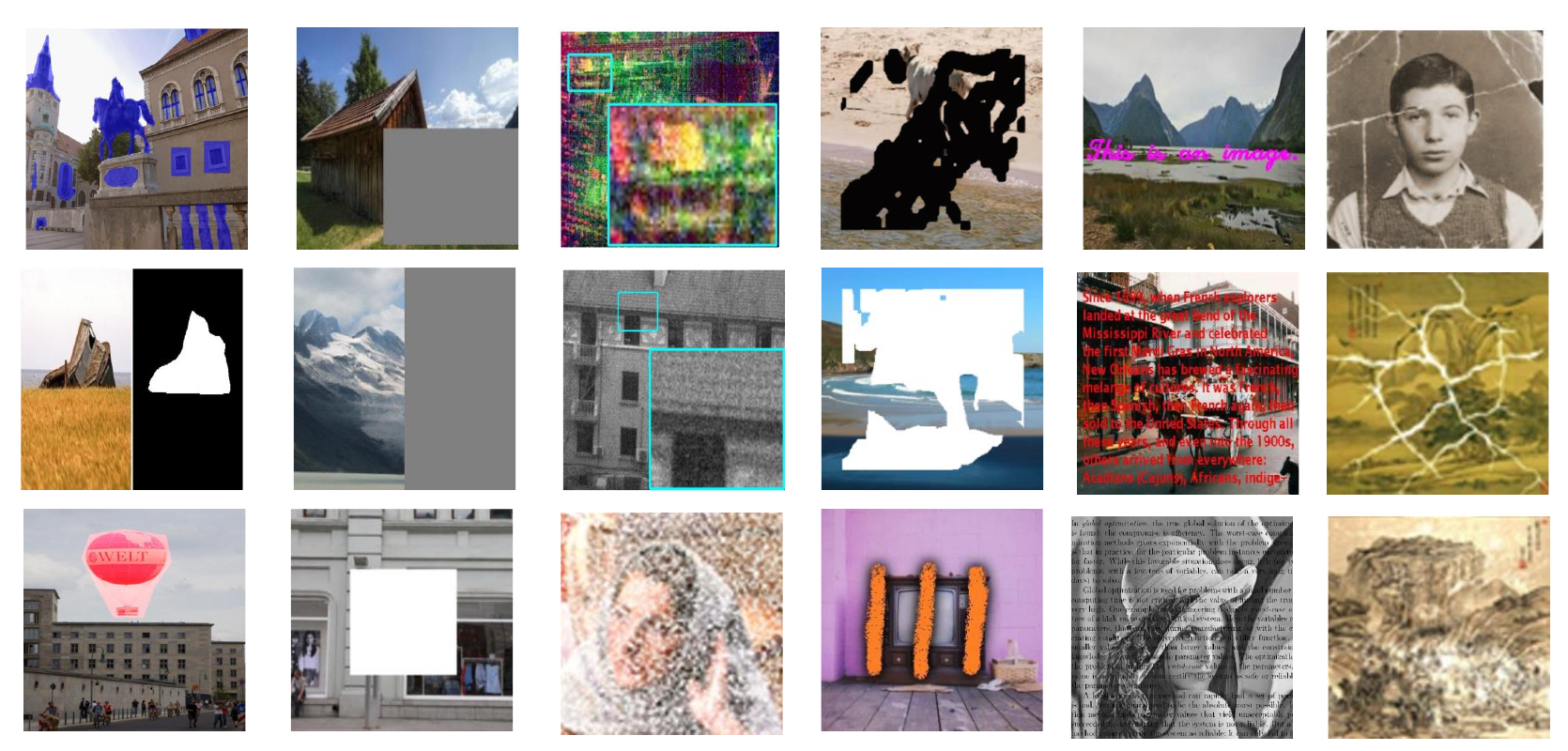}\\
    (a) Object \hspace{1.4cm}(b) Block\hspace{1.4cm}(c) Noise\hspace{1.4cm}(d) Scribble \hspace{1.4cm}(e) Text \hspace{1.4cm}(f) Scratch
  \end{tabular}
  \caption{{Types of masks added to the edited images.}}
\label{typemasks}
\end{figure*}

\subsection{Mask types}

Image inpainting was originally the operation of restoring old images by eradicating scratches and enhancing damaged portions. Presently, it is also employed to eliminate unwanted objects by substituting them with estimated values within the target area. In addition, it is used to repair various distortions or as masks, such text, blocks, noise, scratches, lines, and diverse masks. These masks are used to indicate the areas in an image or video that need to be filled or reconstructed. Figure \ref{typemasks} illustrates the existing distortion or mask types that are used in different image inpainting methods. Several types of masks commonly used in image inpainting are described as follows:

\begin{itemize}
    \item Blocks: a simple mask where a rectangular or a square region in the image is selected for inpainting. It is easy to create and use; however, it may not always be the most accurate representation of the damaged area.
    
    \item Object: in some cases, only specific objects or regions within an image need to be inpainted. Object masks are used to specify these areas for reconstruction while leaving other parts of the image untouched.

    \item Noise: random variations in brightness or color within an image, typically introduced by factors such as low light conditions, sensor limitations, or compression.

    \item Scribble: involve marking the areas to be inpainted with simple strokes or scribbles. These masks provide a rough guideline for the inpainting algorithm and are often used in interactive inpainting systems. These masks are irregularly shaped and can be drawn by hand or generated using algorithms, such as segmentation techniques.

    \item Text: unwanted text overlaid on an image, such as watermarks, captions, or annotations. Inpainting text in an image is the operation of removing or replacing the text with the original content. Text masks are often used in document restoration or editing applications.

    \item Scratches: thin, elongated marks or lines on the sur- face of the image, often caused by physical damage or degradation of the image medium. Generally, this type of mask is used in old pictures.
\end{itemize}

\begin{figure}[t!]
\centering
  \begin{tabular}[b]{c}
    \includegraphics[width=0.81\linewidth]{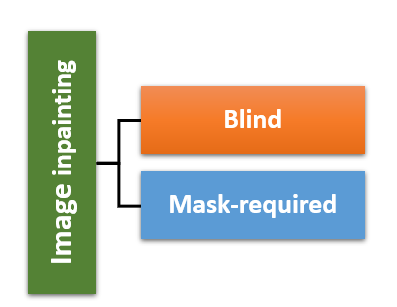}
  \end{tabular}
  \caption{Image inpainting category of methods.}
\label{cat}
\end{figure}

\subsection{Transformer network representations}
To differentiate various types of transformer-based network architectures, we divided the image inpainting models into three categories: blind image inpainting networks, mask-required networks, and GAN-based methods. These categories are illustrated in Figure \ref{cat}, and a description of each method is detailed in Table \ref{tablepara}.

\subsubsection{Blind image inpainting}

The blind image (single-stream) inpainting network is a neural network architecture designed to fill in missing or corrupted regions within an image using only the corrupted image as the input. Through a series of convolutional and/or transformer layers and dedicated inpainting modules, the network predicts the missing regions based on the available information from the input image. Despite its simplicity compared to mask-required (multi-stream) networks, single-stream architectures can still produce impressive inpainting results by learning to infer missing details solely from the provided input image. 
\\
\textbullet   \,   \textbf{CTN} \cite{21-1} introduces the contextual transformer network (CTN) for image inpainting. CTN tackles the challenge of modeling relationships between corrupted and uncorrupted regions while considering their internal structures. Unlike traditional methods, CTN utilizes transformer blocks to capture long-range dependencies and a multi-scale attention module to model intricate connections within different image regions. \\
\textbullet  \, \textbf{ICT} \cite{21-3} combines transformers and CNNs for superior image completion. Transformers capture global structures and various outputs, while CNNs refine textures. This method achieves high fidelity, diverse results.\\
 \labelitemi  \, \textbf{MAT} \cite{21-4} introduces a transformer-based model named the mask-aware transformer (MAT) to efficiently inpaint large holes in high-resolution images. It combines the strengths of transformers for long-range interactions and convolutions for efficient processing. The model incorporates a customized transformer block with a dynamic mask to focus on relevant information and achieve high-fidelity, various image reconstructions. \\
\textbullet  \, \textbf{BAT-Fill} \cite{21-6} is a novel image inpainting method that addresses the limitations of existing CNN-based approaches. While CNNs struggle with capturing long-range features, BAT-Fill leverages a "bidirectional autoregressive transformer" (BAT) for diverse and realistic content generation. Unlike traditional autoregressive transformers, BAT incorporates masked language modeling to analyze context from any direction, enabling better handling of irregularly shaped missing regions. \\
\textbullet \, \textbf{T-former} \cite{22-2}  proposes a new image inpainting method called T-former, aiming to overcome the limitations of CNNs. While CNNs struggle with complex and diverse image damage, T-former leverages a novel attention mechanism inspired by transformers, which offers efficient long-range modeling capabilities while maintaining computational efficiency compared to traditional transformers. \\
\textbullet \, \textbf{PUT} \cite{22-4} is a novel transformer-based method for image inpainting that addresses the information loss issues in existing approaches. Existing methods down-sample images and quantize pixel values, losing information. PUT uses a patch-based autoencoder and an un-quantized transformer that processes the image in patches without down-sampling. The un-quantized transformer directly uses features from the autoencoder, avoiding information loss from quantization.\\
\textbullet  \, \textbf{TransCNN-HAE} \cite{22-6} ] is used for blind image inpainting, which addresses the challenges of unknown and various image damage. Unlike existing two-stage approaches, TransCNN-HAE operates in a single stage that combines transformers and CNNs; it leverages transformers for global context modelling to repair damaged regions and CNNs for local context modelling to reconstruct the repaired image. The crosslayer dissimilarity prompt (CDP) accelerates identifying and inpainting damaged areas.\\
\textbullet  \, \textbf{InstaFormer} \cite{22-7} is a novel network architecture for image-to-image translation that effectively combines global and instance-level information. Global context: utilizes transformers to analyze the overall content of an image, capturing relationships between different parts. (1) Instance-awareness: Incorporates bounding box information to understand individual objects within the image and their interactions with the background. (2) Style control: Enables the application of different artistic styles to the translated image using adaptive instance normalization. (3) Improved instance translation: Introduces a specific loss function to enhance the quality and faithfulness of translated object regions.\\
\textbullet  \, \textbf{Campana et al.} \cite{23-1} Advanced image inpainting has evolved with the integration of transformers in computer vision; however, their high computational costs pose challenges, especially with large, damaged regions. To overcome this, a novel variable-hyperparameter visual transformer architecture is proposed, showing superior performance in reconstructing semantic content, such as human faces. \\
\textbullet \textbf{U2AFN} \cite{23-4} is an uncertainty-aware adaptive feedback network (U2AFN) used to enhance image inpainting for large holes. Unlike conventional methods, U2AFN predicts uncertainty alongside inpainting results and employs an adaptive feedback mechanism. This mechanism progressively refines inpainting regions by utilizing low-uncertainty pixels from previous iterations to guide subsequent learning. \\
\textbullet \textbf{CBNet} \cite{23-8} effective at completing small-sized or specifically masked corruptions, but struggles with large-proportion corrupted images due to limited consideration of semantic relevance. To address this, the authors propose CBNet, a novel image inpainting approach. CBNet combines an adjacent transfer attention (ATA) module in the decoder to preserve contour structure and blend structure–texture information. Additionally, a multi-scale contextual blend (MCB) block assembles multi-stage feature information. Extra deep supervision through a cascaded loss ensures high-quality feature representation.\\
\textbullet \textbf{CoordFill} \cite{23-11} is a novel method using continuous implicit representation to address limitations in image restoration. By utilizing an attentional fast Fourier convolution (FFC)-based parameter generation network, the degraded image is down-sampled and encoded to derive spatial–adaptive parameters. These parameters are then used in a series of multi-layer perceptrons (MLP) to synthesize color values from encoded continuous coordinates. This approach allows the capturing of larger reception fields by encoding high-resolution images at lower resolutions, while continuous position encoding enhances the synthesis of high-frequency textures. Additionally, the framework enables efficient parallel querying of missing pixel coordinates.\\
\textbullet  \, \textbf{CMT} \cite{23-16} is a continuous mask-aware transformer for image inpainting. CMT utilizes a continuous mask to represent error amounts in tokens. It employs masked self-attention with overlapping tokens and updates the mask to model error propagation. Through multiple masked self-attention and mask update layers, CMT predicts initial inpainting results, which are further refined for improved image reconstruction. \\
\textbullet  \, \textbf{TransInpaint} \cite{23-17} is a model for image inpainting that generates realistic content for missing regions while ensuring consistency with the overall context of the image. It utilizes a context-adaptive transformer and a texture enhancement network to produce superior results compared to existing methods.\\
\textbullet  \, \textbf{NDMA} \cite{23-19}
is a lightweight architecture for image inpainting, leveraging nested deformable attention-based transformer layers. These layers efficiently extract contextual information, particularly for facial image inpainting tasks. Comparative evaluations on Celeb HQ and Places2 datasets demonstrate the superiority of the proposed approach.\\
\textbullet  \, \textbf{Blind-Omni-Wav-Net} \cite{23-20} restores corrupted regions without additional mask information. This is challenging due to difficulties in distinguishing between corrupted and valid areas. Existing approaches often struggle to produce plausible results by predicting corrupted regions first. To address this, we propose an end- to-end architecture combining a wavelet query multi-head attention transformer block with omni-dimensional gated attention. The wavelet query multi-head attention provides encoder features using processed wavelet coefficients, while the omni-dimensional gated attention facilitates effective feature transmission from the encoder to decoder. Comparative evaluations on standard datasets demonstrate the superiority of our approach for blind image inpainting through numerical and visual comparisons with state-of-the-art methods.

\begin{landscape}
\begin{table}[]
\def\arraystretch{1.2}
 \scriptsize
\caption{Summarization of crowd counting methods}
\label{tablepara}
\begin{tabular}{|p{1cm}|p{3cm}|c|p{6cm}|p{2cm}|p{1.5cm}|c|c|c|c|c|c|c|}
\hline 
\multirow{5}{*}{\textbf{Network type}} &	\multirow{5}{*}{\textbf{Method}} & \multirow{5}{*}{\textbf{Architecture}} &\multirow{5}{*}{\textbf{Technique}} &\multirow{5}{*}{\textbf{Mask type}} &\multirow{5}{*}{\textbf{Input type} }&\multicolumn{6}{|c|}{\textbf{Dataset}}\\\cline{7-12}

&&&&&&\rotatebox{90}{\textbf{PSV  }} & \rotatebox{90}{\textbf{Places2  }} & \rotatebox{90}{\textbf{CelebA }} & \rotatebox{90}{\textbf{FFHQ  }} & \rotatebox{90}{\textbf{ImageNet }}& \rotatebox{90}{\textbf{ImageNet }} \\ \cline{1-12}

\hline
\multirow{15}{*}{\rotatebox[origin=c]{90}{\textbf{Blind image inpainting}} }
&  CTN \cite{21-1} & Enc-TR-Dec & Stacked-Transformer&Block, Scribble &Image&  \cmark & \cmark & \cmark & \xmark & \xmark& \xmark  \\ \cline{2-12}

& ICT \cite{21-3} & TR-Enc-Dec & Bi-directional transformer&Object, Scribble&Patch & \xmark & \cmark & \xmark & \cmark &\cmark & \xmark 
\\\cline{2-12}

& MAT \cite{21-4} & Conv-TR-Conv & Mask-aware transformer&Block, Scribble&Image& \xmark & \cmark & \cmark & \cmark &\xmark & \xmark  \\
\cline{2-12}
& BAT-Fill \cite{21-6} & TR-Dec & Diver Structure, Texture generation networks&Scribble& Image &\cmark & \cmark & \cmark & \xmark &\xmark & \xmark \\
\cline{2-12}
& T-former \cite{22-2} & TR & Transformer Unet &Scribble&Image &
\cmark & \cmark & \cmark & \xmark &\xmark & \xmark\\
\cline{2-12}

& PUT & ENc-TR-Dec & Un-Quantized Transformer (UQ-Transformer) &Scribble&Patch& \xmark & \cmark & \xmark & \cmark &\cmark & \xmark\\
\cline{2-12}

& TransCNN-HAE \cite{22-6} & Tr-Deco &  Transformer-CNN Hybrid AutoEncoder &Scribble&Patch &\cmark & \cmark & \cmark & \cmark &\xmark & \xmark\\
\cline{2-12}
& InstaFormer \cite{22-7} & Enc-TR-Dec & ViT block &Obejct&Image &\xmark & \xmark & \xmark & \xmark &\xmark & \cmark
\\
\cline{2-12}

& Campana et al. \cite{23-1} & Enc-TR-Dec &Cariable-hyperparameter visual transformer &Scribble &Image&\cmark & \cmark & \cmark & \xmark &\xmark & \xmark\\\cline{2-12}
& U2AFN \cite{23-4}  &Enc-TR-Dec  & Uncertainty-aware adaptive feedback network&Block, Scribble& Image&\cmark & \cmark & \cmark & \xmark &\xmark & \xmark\\\cline{2-12}

&CBNet \cite{23-8}& Enc-TR-Dec & Cascading blend network &Scribble &Image& \cmark & \cmark & \cmark & \xmark &\xmark & \xmark\\\cline{2-12}

&CoordFill \cite{23-11}& Enc-TR & Attentional Fast Fourier Conv (FFC), multi-layer perceptrons (MLP)& Object, Block&Image & \xmark & \cmark & \cmark & \xmark & \xmark & \xmark\\\cline{2-12}

&CMT \cite{23-16}& TR & Self-attention and mask update (MSAU) layers &Scribble, scratch &Image&\xmark & \cmark & \cmark & \xmark &\xmark & \xmark\\\cline{2-12}

&TransInpaint \cite{23-17}  & TR-Enc-Dec & DETR, context-adaptive transformer&Block, Scribble &Image &\xmark & \cmark & \cmark & \cmark & \cmark & \xmark%
\\\cline{2-12}

&NDMA \cite{23-19}&  Enc-Dec-Enc-Att& Deformable attention-based transformer &Scribble &Face image &\xmark & \xmark & \cmark & \xmark &\xmark & \xmark\\\cline{2-12}
&Blind-Omni-Wav-Net \cite{23-20}& TR  & Wavelet query multi-head attention (WQMA), omnidimensional gated attention (OGA) & Scribble,  Scratch&Image &\cmark & \cmark & \cmark & \cmark &\xmark & \xmark\\\cline{2-12}

\hline

\multirow{8}{*}{\rotatebox[origin=c]{90}{\textbf{Mask-required }} }
&  ZITS \cite{21-5} &Enc-TR-Dec &Consecutive encoder decoder networks with transfromer TR&Scribble&Four image& \xmark & \cmark & \cmark & \xmark & \xmark & \cmark
\\
\cline{2-12}
&APT \cite{22-1} & TR &Atrous Pyramid Transformer &Scribble &Two images&\cmark & \cmark & \cmark & \xmark &\xmark & \xmark\\
\cline{2-12}
& SPN \cite{23-2} & Enc-TR-Dec&Semantic Pyramid Networ  &Scribble&Image&\cmark & \cmark & \cmark & \xmark &\xmark & \xmark\\
\cline{2-12}
& SWMH \cite{23-6} & Enc-TR-Dec & SWMH transformer blocks &Object, Scribble&Mask, Image &\xmark & \cmark & \cmark & \xmark &\xmark & \xmark\\\cline{2-12}

&ZITS++ \cite{23-7} &Enc-TR-Dec &Transformer Structure Restorer (TSR) module for holistic structural priors at low resolution&Scribble &Four Image&\xmark & \cmark & \xmark & \cmark &\xmark & \xmark\\
\cline{2-12}
&TransRef \cite{23-9}& TR & Reference-patch alignment (Ref-PA) module,reference-patch transformer (Ref-PT) module  &Object, Scribble&Image &\xmark & \xmark & \xmark & \xmark &\xmark & \cmark
\\\cline{2-12}

\hline
\multirow{10}{*}{\rotatebox[origin=c]{90}{\textbf{GAN-based}} }
&  ACCP-GAN \cite{21-2} &Unet-Enc-TR-Dec &Consecutive networks with TR& Scribble &Image& \cmark & \cmark & \cmark & \xmark & \xmark & \cmark
\\\cline{2-12}
& AOT-GAN & Enc-TR-Dec & Aggregated Contextual-Transformation &Scribble&Image&\xmark & \cmark & \cmark & \xmark &\xmark & \xmark\\
\cline{2-12}
& HiMFR \cite{22-5} & GAN-TR & Vision Transformer &Object&Image&\xmark & \xmark & \cmark & \xmark &\xmark & \cmark
\\
\cline{2-12}
& Wang et al.\cite{22-6}  & Enc-Tr-Dec & Deep semantic structure modeling module (U-CITB)&Scribble&Image&\cmark & \cmark & \cmark & \cmark &\xmark & \xmark\\
\cline{2-12}
&Li et al. \cite{23-3}& Enc-TR-Dec & Prior-Driven Fused Contextual Transformation Network &Scribble&Image &\cmark & \cmark & \cmark & \xmark &\xmark & \xmark\\
\cline{2-12}
& Swin-GAN \cite{23-5} & Enc-Att-Dec & Transformer-based Descriminator&Scribble&Image&\cmark & \cmark & \cmark & \xmark &\xmark & \xmark\\\cline{2-12}
&SFI-Swin \cite{23-10}  & TR & Swin-Tr& Block, Scribble &Face image&\xmark & \xmark & \cmark & \xmark &\xmark & \xmark\\
\cline{2-12}
&IIN-GCMAM \cite{23-11}  & Enc-(Dec+Atts) & CNN Encoder +multi-level attention mechanism decoder (MAM-decoder) & Scribble&Image &\cmark & \xmark & \cmark & \xmark &\xmark & \xmark\\ \cline{2-12}
&WAT-GAN \cite{23-13}& Enc-TR-Dec & Window Aggregation Transformer (WAT), GAN & Scribble&Image&\xmark & \cmark & \cmark & \xmark &\xmark & \xmark\\ \cline{2-12}
&UFFC \cite{23-14}& - & Unbiased Fast Fourier Convolution (UFFC)  &Block, Scribble &Image&\xmark & \cmark & \cmark & \xmark &\xmark & \xmark\\\cline{2-12}

&PATMAT \cite{23-15}& GAN-TR  & Mask-Aware Transformer
(MAT)  &Object &Face image &\xmark & \xmark & \cmark & \xmark &\xmark & \xmark\\ \cline{2-12}
&GCAM \cite{23-18} & Unet-TR  & Lightweight Attention using Group Convolution module (LAGC) &Block, Scribble&Image&\cmark & \cmark & \cmark & \xmark &\xmark & \xmark\\ 
 \bottomrule
\end{tabular}
\end{table}
\end{landscape}

\subsubsection{Mask-required image inpainting}

Mask-required networks for inpainting have a neural network architecture that utilizes multiple input streams to perform the inpainting task. The mask is fed into the network with the distorted image. This architecture is designed to handle various types of input information, which can improve the inpainting performance by leveraging different features and representations. The network takes in multiple streams of input data, each representing different types of information relevant to the inpainting task. For example, one stream could contain the corrupted image, another stream could contain additional contextual information, such as edge maps or semantic segmentation masks, and another stream could contain guidance from reference images.\\
\textbullet  \, \textbf{ZITS } \cite{21-5} tackles the challenge of restoring both textures and structures in corrupted images. While CNNs struggle with capturing holistic structures, attention-based models are computationally expensive for large images. For that, a structure restorer network uses a transformer model in a low-resolution space to efficiently recover the overall structure of the image. The recovered structure is then integrated with existing inpainting models to add details and textures.\\
\textbullet  \, \textbf{APT} \cite{22-1} is a two-stage image inpainting framework using a novel "atrous pyramid transformer" (APT). APT captures long-range dependencies to reconstruct damaged areas, while a "dual spectral transform convolution" (DSTC) module refines textures. \\
\textbullet  \, \textbf{SPN} \cite{23-2} restoring realistic content in images with missing regions is challenging. Existing image inpainting models often produce blurred textures or distorted structures in complex scenes due to contextual ambiguity. To address this, we propose the semantic pyramid network (SPN), leveraging multi-scale semantic priors learned from pretext tasks. SPN comprises two components: a prior learner distilling semantic priors into a multi-scale feature pyramid, ensuring a coherent understanding of global context and local structures, and a fully context-aware image generator progressively refining visual representations with the prior pyramid. Optionally, variational inference enables probabilistic inpainting.\\
\textbullet  \, \textbf{SWMH } \cite{23-6} is a model that combines a specialized transformer, named the stripe window multi-head (SWMH) transformer, with a traditional CNN. It has a novel loss function to enhance color details beyond RGB channels. \\

 \textbullet  \, \textbf{ZITS++} \cite{23-7} is an improved model of the authors previous work, ZITS. ZITS++ combines a specialized transformer with a traditional CNN. It introduces the transformer structure restorer (TSR) module for holistic structural priors at low resolution, upsampled by the simple structure upsampler (SSU). Texture details are restored using the Fourier CNN texture restoration (FTR) module, enhanced by Fourier and large-kernel attention convolutions. The upsampled structural priors from TSR are further processed by the structure feature encoder (SFE) and optimized incrementally with the zero-initialized residual addition (ZeroRA). Additionally, a new masking positional encoding addresses large, irregular masks.\\
 \textbullet  \, \textbf{TransRef} \cite{23-9} is a transformer-based encoder–decoder network for reference-guided image inpainting. The guidance process involves progressively aligning and fusing referencing features with the features of the corrupted image. To precisely utilize reference features, they introduce the reference-patch alignment (Ref-PA) module, which aligns patch features from both reference and corrupted images while harmonizing style differences. Additionally, the reference-patch transformer (Ref-PT) module refines the embedded reference feature. 
 \textbullet  \, \textbf{UFFC} \cite{23-14} examines the limitations of using the vanilla FFC module in image inpainting, including spectrum shifting and limited receptive fields. To address these issues, a novel unbiased fast Fourier convolution (UFFC) module was proposed, incorporating range transform, absolute position embedding, dynamic skip connection, and adaptive clipping. The experimental results demonstrate that the UFFC module outperforms existing methods in capturing texture and achieving faithful reconstruction in image inpainting tasks.

\begin{figure*}[t!]
\centering
  \begin{tabular}[b]{c}
    \includegraphics[width=1\linewidth]{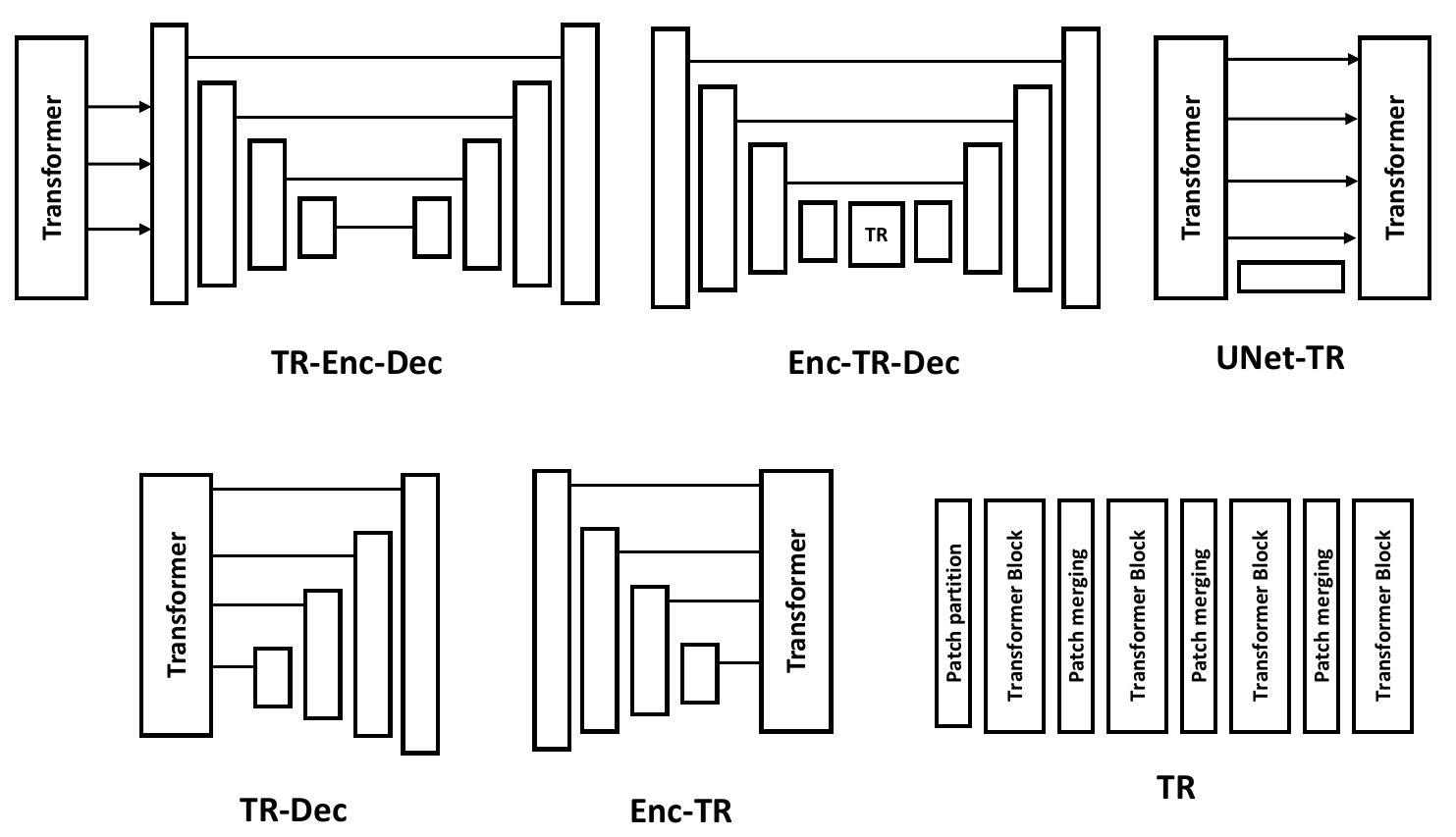}
  \end{tabular}
  \caption{{Transformer -based architectures used in different image inpainting methods. The transformer blocks are combined with different convolutional neural network (CNN)-based parts including encoder–decoders such as in Enc-TR-Dec, TR-Dec, and Enc-Tr representations. Some architectures used a pure transformer-based network, such as TR and UNet-TR, which utilizes a transformer-based encoder and decoder. }}
\label{typeoftr}
\end{figure*}
\subsubsection{GAN With Transformer image inpainting:}

GAN-based image inpainting utilizes GANs to fill in missing or corrupted regions of an image. In this approach, two neural networks are trained simultaneously: a generator network and a discriminator network. The generator generates realistic content for the missing regions, while the discriminator tries to distinguish between the inpainted images and real images. Through adversarial training, the generator learns to produce convincing inpainted images that fool the discriminator. This method often produces visually appealing results by leveraging the adversarial loss to capture high-level image structures and textures. However, GAN-based inpainting is prone to issues such as mode collapse or blurriness, requiring careful optimization and architectural choices to address these challenges.\\
\textbullet  \, \textbf{ACCP-GAN} \cite{21-2}is a method for automatically repairing defects in serial section images used in histology studies. ACCP-GAN combines two stages: one to detect and roughly fix damaged areas, and another to precisely refine the repairs. The model leverages transformers and convolutions to analyze neighboring images and healthy regions within the defective image, achieving high accuracy in both segmentation and restoration tasks.\\ 
\textbullet  \, \textbf{AOT-GAN} \cite{22-3} tackles challenges in high-resolution image inpainting. It improves both context reasoning and texture synthesis, leading to more realistic reconstructions compared to existing approaches. This method is particularly effective for large, irregular missing regions.\\
\textbullet  \, \textbf{HiMFR}\cite{22-5} is a system for recognizing masked faces. HiMFR first detects masked faces using a pre- trained model and inpaints the occluded regions using a GAN-based method. Finally, it recognizes the face, masked or reconstructed, using a hybrid recognition module. Experiments show competitive performance on benchmark datasets.\\
\textbullet  \, \textbf{Wnag et al.}\cite{22-6} addresses limitations in reconstructing large, damaged areas with image inpainting. The authors propose enhanced gated convolution to extract detailed features from the masked region using a gating mechanism. U-net-like deep structure modeling combines transformers’ long-range modeling with CNNs’ texture learning to capture global structures. Next, the reconstruction module merges shallow and deep features to generate the final inpainted image.\\
\textbullet  \, \textbf{Li et al.} \cite{23-3} recent image inpainting advancements perform well on simple backgrounds but struggle with complex images due to the lack of semantic understanding and distant context. To address this, a semantic prior-driven fused contextual transformation network was proposed. It utilizes a semantic prior generator to map features from ground truth and damaged images, followed by a fusion strategy to enhance multi-scale texture features and an attention aware module for structure restoration. Additionally, a mask-guided discriminator improves output quality. Results on various datasets show significant improvements over existing methods. \\
\textbullet  \, \textbf{Swin-GAN} \cite{23-5}  presents a transformer-based method for image inpainting, aiming to overcome limitations in capturing global and semantic information. The technique utilizes self-supervised attention and a hierarchical Swin transformer in the discriminator. Experimental results show superior performance compared to existing approaches, demonstrating the effectiveness of the proposed transformer-based approach.\\
\textbullet  \, \textbf{SFI-Swin} \cite{23-10} image inpainting involves filling in the holes or missing parts of an image. When it comes to inpainting face images with symmetric characteristics, the challenge is even greater than for natural scenes. Existing powerful models struggle to fill in missing parts while considering both symmetry and homogeneity. Additionally, standard metrics for assessing repaired face image quality fail to capture the preservation of symmetry between rebuilt and existing facial features. To address this, the authors propose a GAN-transformer-based solution: multiple discriminators that independently verify the reality of each facial organ, combined with a transformer-based network. They also introduce a novel metric called the "symmetry concentration score" to measure the symmetry of repaired face images.\\
\textbullet  \, \textbf{IIN-GCMAM} \cite{23-11} is an image inpainting network using gated convolution and a multi-level attention mechanism to address deficiencies in existing methods. By weighing features with gated convolutions and employing multi-level attention, it enhances global structure consistency and repair result precision. Extensive experiments on datasets, such as Paris Street View (PSV) and CelebA, have validated its effectiveness.\\
\textbullet  \, \textbf{WAT-GAN} \cite{23-13} is a novel transformer network with cross-window aggregated attention use to address limitations of convolutional networks, such as over-smoothing and limited long-range dependencies. Integrated into a generative adversarial network model, this approach embeds the window aggregation transformer (WAT) module to enhance information aggregation between windows without increasing computational complexity. Initially, the encoder extracts multi-scale features using convolution kernels of varying scales. These features are then input into the WAT module for inter-window aggregation, followed by reconstruction by the decoder. The resulting image undergoes assessment by a global discriminator for authenticity. Experimental validation demonstrates that the transformer window attention network enhances the structured texture of restored images, particularly in scenarios involving large or complex structural restoration tasks.\\
\textbullet  \, \textbf{PATMAT} \cite{23-15} is a method for face inpainting that enhances the preservation of facial details and identity. By fine-tuning a MAT with reference images, it outperforms existing models in quality and identity preservation.\\
\textbullet  \, \textbf{GCAM} \cite{23-18} is a lightweight image inpainting method that emphasizes both restoration quality and efficiency on limited processing platforms. By combining group convolution and a rotating attention mechanism, the traditional convolution module is enhanced or replaced. Group convolution enables multi-level inpainting, while the rotating attention mechanism addresses information mobility issues between channels. A parallel discriminator structure ensures local and global consistency in the inpainting process. Experimental results show that the proposed method achieves high-quality inpainting while significantly reducing inference time and resource usage compared to other lightweight approaches.


\section{Video inpainting}

Video inpainting using transformers is a technique that uses visual transformer models to fill in missing or corrupted parts of a video sequence. By utilizing the transformer's ability to capture long-range dependencies in the data, this approach aims to seamlessly reconstruct the missing regions in video frames based on the surrounding context. Many method have been proposed. For that, in this section a description of each one these method will be described.

\textbullet  \, \textbf{FuseFormer} \cite{v21-1} is a transformer model tailored for video inpainting tasks to address issues with blurry edges. It utilizes Soft Split and Soft Composition operations to enhance fine-grained feature fusion. Soft Split divides feature maps into patches with overlap, while Soft Composition stitches patches together, allowing for more effective interaction between neighboring patches. These operations are integrated into tokenization and de-tokenization processes for better feature propagation. Additionally, FuseFormer enhances the capability of 1D linear layers to model 2D structures, improving sub-patch level feature fusion. The evaluation results demonstrated the superiority of FuseFormer over existing methods in both quantitative and qualitative assessments.\\
\textbullet  \, \textbf{FAST} \cite{v21-2} is a frequency-aware spatiotemporal transformer used for video inpainting detection. It utilizes global self-attention mechanisms to capture long-range relations and employs a spatiotemporal transformer framework to detect spatial and temporal connections. Additionally, FAST exploits frequency domain information using a specially designed decoder. Experimental results show competitive performance and good generalization. \\
\textbullet  \, \textbf{DSTT} \cite{v21-3} is a decoupled spatial–temporal trans- former used for efficient video inpainting. It separates learning spatial–temporal attention into two tasks: one for temporal object movements and another for background textures. This allows precise inpainting. Additionally, a hierarchical encoder is used for robust feature learning.\\
\textbullet  \, \textbf{E2FGVI } \cite{v22-1} is an end-to-end framework for flow-guided video inpainting to improve the efficiency and effectiveness compared to existing methods. It replaces separate hand-crafted processes with three trainable modules: flow completion, feature propagation, and content hallucination. These modules correspond to previous stages but can be jointly optimized, leading to better results.\\ 
\textbullet  \, \textbf{FGT } \cite{v22-2} is a flow-guided transformer used for high- fidelity video inpainting, which utilizes motion discrepancy from optical flows to guide attention retrieval in transformers. A flow completion network is introduced to restore corrupted flows by leveraging relevant flow features within a local temporal window. With completed flows, the content is propagated across frames and flow-guided transformers are employed to fill in corrupted regions. Transformers are decoupled along temporal and spatial dimensions to integrate the completed flows for spatial attention. Additionally, a flow-reweight module controls the impact of completed flows on each spatial transformer. For efficiency, a window partition strategy is employed in both the spatial and temporal transformers. \\
\textbullet  \, \textbf{DeViT} \cite{v22-2} ], deformed vision transformer (DeViT), presents three key innovations: DePtH for patch alignment, MPPA for enhanced feature matching, and STA for accurate attention assignment. DeViT outperforms previous methods in quality and quantity, setting a new state-of-the-art for video inpainting.\\
\textbullet  \, \textbf{DLFormer} \cite{v22-4} is a discrete latent transformer. Unlike previous methods operating in continuous feature spaces, DLFormer utilizes a discrete latent space, leveraging a compact codebook and autoencoder to represent the target video. By inferring proper codes for unknown areas via self-attention, DLFormer produces fine-grained content with long-term spatial–temporal consistency. Additionally, it enforces short-term consistency to reduce temporal visual jitters. \\
\begin{figure*}[t!]
\centering
  \begin{tabular}[b]{c}
    \includegraphics[width=.45\linewidth]{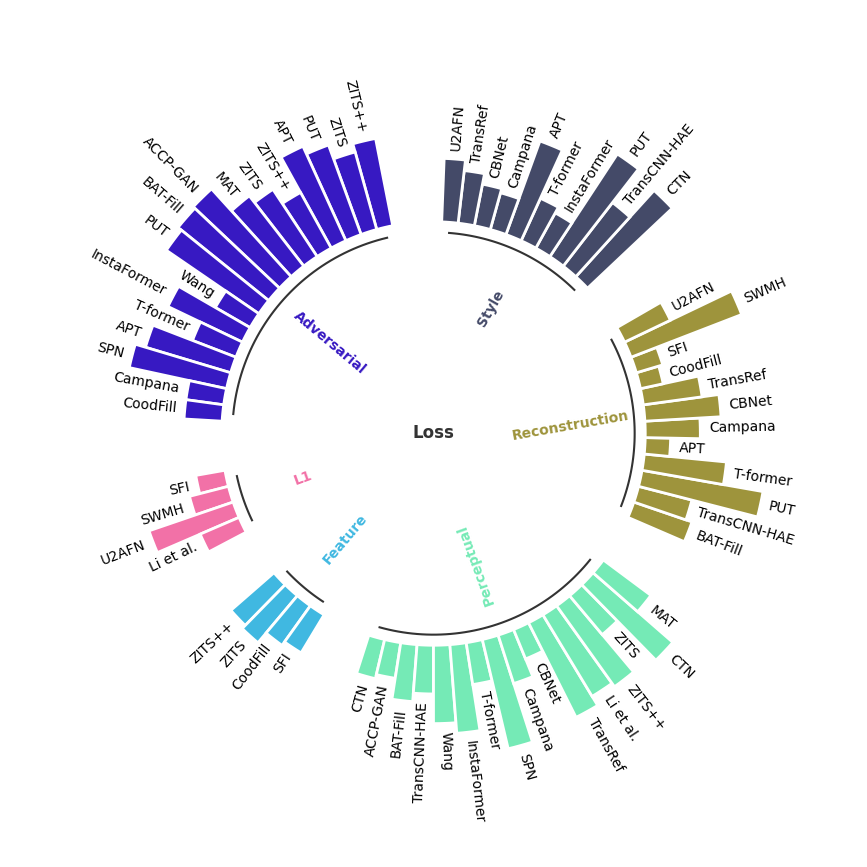}\\
    (a) Image inpainting
  \end{tabular}
    \begin{tabular}[b]{c}
    \includegraphics[width=.45\linewidth]{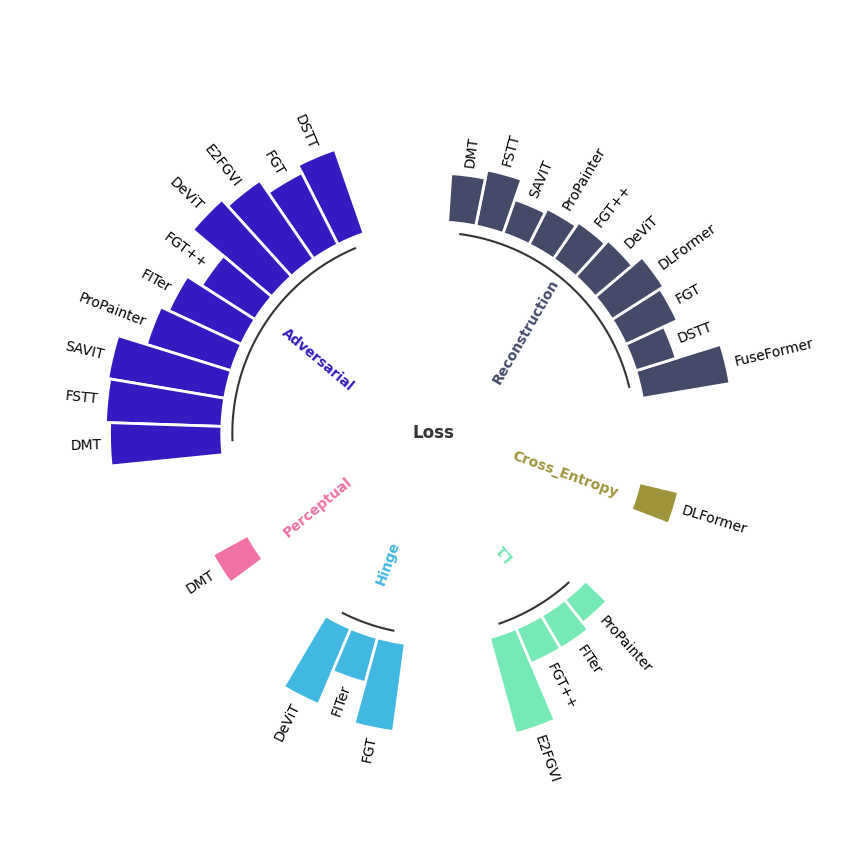}\\
    (a) Video inpainting
  \end{tabular}
  \caption{{Loss functions used in each transformer-based image or video inpainting method. Five loss function are combined for transformer-based learning. }}
\label{loss}
\end{figure*}
\textbullet  \, \textbf{DMT} \cite{v23-1} is a dual-modality-compatible inpainting framework used to address deficiencies in video inpainting. DMT\_img, a pretrained image inpainting model, serves as a prior for distilling DMT\_vid, enhancing performance in deficiency cases. The self-attention module selectively incorporates spatiotemporal tokens, accelerating inference and removing noise signals. Additionally, a receptive field contextualizer improves performance further. \\
\textbullet  \, \textbf{FGT++} \cite{v23-2} is an enhanced version of the flow-guided transformer (FGT), resulting in more effective and efficient video inpainting. FGT++ addresses query degradation using a lightweight flow completion network and introduces flow guidance feature integration and flow-guided feature propagation modules. The transformer is decoupled along the temporal and spatial dimensions, utilizing flows for token selection and employing a dual-perspective multi-head self-attention (MHSA) mechanism. Experimental results show that FGT++ outperforms existing video inpainting networks in both quality and efficiency.\\
\textbullet  \, \textbf{} Liao et al., \cite{v23-3} propose an automatic video inpainting algorithm for clear street views in autonomous driving. Using depth/point cloud guidance, this method removes traffic agents from videos and fills missing regions. By creating a dense 3D map from point clouds, frames are geometrically correlated, allowing for straightforward pixel transformation. Multiple videos can be fused through 3D point cloud registration, addressing long-time occlusion challenges.\\
\textbullet  \, \textbf{FITer} \cite{v23-4} is a video inpainting method that enhances missing region representations using a feature pre-inpainting network (FPNet) before the transformer stage. This improves the accuracy of self-attention weights and dependency learning. FITer also employs an interleaving transformer with global and window-based local self-attention mechanisms for efficient aggregation of spatial– temporal features into missing regions.\\
\textbullet  \, \textbf{ProPainter} \cite{v23-5} is an enhanced framework for video inpainting that addresses the limitations in flow-based propagation and spatiotemporal transformers. ProPainter combines image and feature warping for more reliable global correspondence and employs a mask-guided sparse video transformer for increased efficiency. \\
\textbullet  \, \textbf{SViT} \cite{v23-6} is a new transformer-based video inpainting technique that leverages semantic information to enhance reconstruction quality. Using a mixture-of-experts scheme, multiple experts can be trained to handle mixed scenes with various semantics. By producing different local network parameters at the token level, this method achieves semantic-aware inpainting results \\
\textbullet  \, \textbf{FSTT} \cite{v23-7} uses a flow-guided spatial temporal transformer (FSTT) for video inpainting, which effectively utilizes optical flow to establish correspondence between missing and valid regions in spatial and temporal dimensions. FSTT incorporates a flow-guided fusion feed-forward module to enhance features with optical flow guidance, reducing inaccuracies during MHSA. Additionally, a decomposed spatiotemporal MHSA module captures dependencies effectively. To improve efficiency, a global–local temporal MHSA module was designed.

\section{\textbf{Loss Functions:}}

Following a review of the cited transformer-based methods for image or video inpainting, various loss functions have been utilized to guide the generation of realistic results. Generally, to train these methods, authors combined more than one loss function in their implementation due to the difference in the objective of each one of these functions. The most used loss functions in image inpainting include mean absolute error loss (L1)  \cite{L1}, Adversarial Loss, Perceptual Loss \cite{perceptual}, Reconstruction Loss \cite{reconstruction}, Style Loss \cite{style}, and Feature Map Loss. Also some other loss functions are used in a small number of papers like Mask, SSIM loss function used in ACCP-GAN \cite{21-2}, binary cross entropy loss \cite{bcross}, Cross entropy loss \cite{cross}, and diversified Markov random field loss \cite{MRF}. 

]. For image inpainting and image translation tasks, including image generation and image segmentation, incorporating various loss functions can produce better results visually and is semantically effective. 
These loss functions can be categorized into three classes: contextual-based, style-based, and structure-based loss. Contextual-based loss functions focus on preserving the content or semantic information of the image, ensuring that the inpainted regions are coherent and homogeneous with the neighboring regions. Furthermore, they can be used to measure the similarity between the inpainted image and the ground truth in terms of both low-level details and high-level structures, preserving realistic content. For this category, L1 and reconstruction loss functions can be found \cite{L1,reconstruction}.The style-based loss category is focused on capturing high-level semantic information rather than pixel-level details. It specifically targets the texture and artistic style of the original image, which is achieved by comparing the statistics of feature maps across different layers of the network. In this category exists perceptual loss, style loss and adversarial loss \cite{style}. Structural loss that can be categorized as a type of contextual loss, which emphasizes maintaining the contextual coherence and structural integrity of the inpainted image, preserving the surrounding content. Figure \ref{loss} summarizes the used loss functions in this review. By the following a description of each one of the used loss functions.\\
\textbf{Mean Absolute Error (L1) loss}measures the absolute pixel-wise differences between the inpainted image and the ground-truth. It allows the difference between the generated image to be minimized to be close to the original image in terms of pixel values. \\
\textbf{Adversarial loss} introduced for GANs, consists of a generator and a discriminator. The generator aims to produce realistic images, while the discriminator learns to distinguish between real and generated images. The adversarial loss encourages the generator to generate visually robust content.\\
\textbf{Perceptual loss} or High Receptive Field (HRF), focuses
on capturing high-level semantic information. It defines the difference between original input features and reconstructed features. By minimizing perceptual loss, the inpainted image is encouraged to capture the structural similarities with the original image.\\
\textbf{Reconstruction loss} evaluates the comparison between the inpainted and the original image after each transformation. It also assesses the definite differences between the generated images and real images.\\
\textbf{Style loss} captures the texture and artistic style of the original image. It is computed by comparing the statistics of feature maps across different layers. By minimizing style loss, the inpainted regions are encouraged to mimic the artistic style of the surrounding image content.\\
\textbf{Feature Map loss}measures the similarity between feature maps extracted from the inpainted image and those from the ground-truth image. It encourages the inpainted regions to preserve important visual structures and textures present in the original image. Feature map loss is often used in conjunction with perceptual loss to guide the inpainting process effectively.\\
\textbf{Hinge loss} for video inpainting is used in adversarial settings. which can help in training discriminators to distinguish between real and inpainted video frames, ensuring that inpainted images are indistinguishable from the original content. It is not commonly used for the direct inpainting tasks but can improve the quality of inpainted videos in GAN-based approaches.\\
\textbf{Cross-entropy loss} in the context of video inpainting, is
mainly suitable for classification tasks within the inpainting process, such as segmenting regions to be inpainted. It measures the difference between the predicted and actual distribution of classes (e.g., inpainted vs. not inpainted pixels).

\section{Image and video inpainting datasets}
To evaluate the image inpainting method, various datasets have been used.  Paris Street View dataset \cite{psv} was created for image inpainting, and others are from other tasks, such as Places2 \cite{places} for scenes recognition, CelebA-HQ \cite{celeba} and FFHQ \cite{FFHQ} for Face recognition, Youtube-VOS \cite{youtube} for video object segmentation. In this section, the most frequent datasets used in the transformed-based image inpainting methods are reviewed.

\textbf{ Paris Street View Dataset:}
The Paris Street View dataset \cite{psv} consists of 14,900 training images and 100 test images captured from street views in Paris. These images primarily focus on the city’s buildings, making the dataset valuable for tasks related to urban scenes and architectural elements.

\textbf{CelebA-HQ Dataset:} CelebA-HQ \cite{celeba} is an extension of the CelebA dataset, providing high-quality images of celebrities with diverse attributes. It contains over 30,000 high-resolution images (1024 × 1024 pixels) of celebrities in various poses, lighting conditions, and backgrounds. CelebA-HQ is commonly used for tasks such as facial recognition, attribute classification, and image generation, including image inpainting.

\textbf{Places2 Dataset:}
Places2 \cite{places} is a large-scale dataset focusing on scene understanding, containing images of various indoor and outdoor scenes from around the world. It includes over 10 million images covering 365 scene categories, ranging from natural landscapes to urban environments. Places2 is used for several tasks such as scene classification, semantic segmentation, and image inpainting.

\textbf{ FFHQ  Dataset:}
Flickr-Faces-HQ (FFHQ) \cite{FFHQ} dataset is a high-quality image collection of human faces: 70,000 high-quality images with a resolution of 1024 × 1024 pixels. The dataset contains various images with variations in terms of age, ethnicity, and image background. In addition, it includes a diverse range of attributes such as eyeglasses, sunglasses, and hats. FFHQ is used in different tasks, such as image generation, super-resolution, denoising, and inpainting

\textbf{YouTube-VOS Dataset:}
The YouTube-VOS (Video Object Segmentation) \cite{youtube} dataset is designed for the task of semi-supervised video object segmentation, where the goal is to segment objects of interest in videos. It contains high-resolution video sequences with pixel-level annotations for foreground objects across multiple frames. The YouTube-VOS dataset is used for video object segmentation and video inpainting tasks.

\textbf{DAVIS Dataset:}
The Densely Annotated Video Segmentation (DAVIS) dataset is a comprehensive resource designed specifically for the task of video object segmentation, offering high-quality annotations across consecutive frames for precise delineation of object boundaries. It contains 50 high-quality video sequences. Furthermore, DAVIS provides a benchmark for evaluating algorithms in the field of video segmentation, in addition to video inpainting.


\section{Results and discussion}

\subsection{\textbf{Evaluation Metrics }}
To evaluate the performance of image or video inpainting methods, a set of metrics are used to compare between the generated image and the ground-truth. In this section, we selected the most used metrics for image and video inpainting, including peak signal-to-noise ratio (PSNR), structural similarity index (SSIM), learned perceptual image patch similarity (LPIPS), and Frechet inception distance (FID).
The metrics can be divided into two categories: pixel-based metric to evaluate the quality of the generated images and patch-based metric to compute the perceptual similarity between two images. FID and LPIPS are patch-based metrics, and PSNR and SSIM are pixel-based metrics.

\subsection{Pixel-based metrics:}
Pixel-based metrics evaluate images at the level of individual pixels. The PSNR and SSIM are pixel-based metrics used to evaluate image inpainting method.

\textbf{Peak Signal-to-Noise Ratio (PSNR ):} Used to evaluate the quality of the generated image by comparing it with the ground-truth. A higher PSNR indicates less noise and better quality. The PSNR is defined as follows:
\begin{equation}
    PSNR = 10 \log_{10} \left( \frac{MAX_I^2}{MSE} \right)
\end{equation}
Where $MAX_I$ is the maximum pixel value of the image, and 
$MSE$ is the Mean Squared Error between the ground-truth and generated image.

\textbf{Structural Similarity Index (SSIM):} This metric compares how similar two images are in terms of luminance, contrast, and structure, mimicking human perception. A higher SSIM indicates that the images are more alike. The SSIM is defined as follows:
\begin{equation}
   SSIM(x, y) = \frac{(2 \mu_x \mu_y + C_1)(2 \sigma_{xy} + C_2)}{(\mu_x^2 + \mu_y^2 + C_1)(\sigma_x^2 + \sigma_y^2 + C_2)} 
\end{equation}

Where $\mu_x$ and $\mu_y$ denote the mean luminance values of images $x$ and $y$ respectively. the  $\sigma_x$ and $\sigma_y$ denote the standard deviations of $x$ and $y$ respectively. And $\sigma_{xy}$ is the covariance between $x$ and $y$. While $C_1$ and $C_2$ are constant.

\subsection{Patch-based metrics:} 
Patch-based metrics evaluate images by comparing patches or local regions instead of individual pixels. These metrics typically use deep learning techniques to extract features from the patches. LPIPS and FID are patch-based metrics used to evaluate image inpainting methods.

\textbf{Learned Perceptual Image Patch Similarity (LPIPS):}
This metric is used to evaluate the perceptual similarity between two images. Unlike traditional metrics, such as the MSE or PSNR, which measure pixel-wise differences, LPIPS is designed to capture perceptual differences that align more closely with human perception. LPIPS calculates the average Euclidean distance between the feature representations of corresponding patches or layers extracted from the images. This distance reflects the perceptual difference between the two images. The LPIPS metric can be expressed as follows: 
\begin{equation}
\text{LPIPS}(I_1, I_2) = \frac{1}{N} \sum_{i=1}^{N} \| \phi(I_1)_i - \phi(I_2)_i \|_2
\end{equation}

 Where $I_1$ and $I_2$ represent the input image and the ground-truth. $\phi(I_1)$ denotes the feature representation of image $I$. $\phi(I)$  represents the feature vector of the $i-th$ patch or layer in the feature representation of image $I$. $N$ is the total number of patches or layers in the feature representation. $\|.\|_2$ denotes the Euclidean distance.
 
\textbf{Fréchet Inception Distance (FID):} This is used to evaluate the quality of generated images in generative models such as those using GANs. This metric is considered as to be patch-based metrics in some papers \cite{23-10}, while in other, it is considered to be a features-based metric. A lower FID value signifies a higher consistency between the two image sets. For video inpainting, researchers used VFID. The FID is formulated as follows: 

\begin{equation}
FID = \lVert \mu_{\text{real}} - \mu_{\text{gen}} \rVert^2 + \text{Tr}(\Sigma_{\text{real}} + \Sigma_{\text{gen}} - 2(\Sigma_{\text{real}}\Sigma_{\text{gen}})^{0.5})
\end{equation}

 Where $\|.\|_2$ denotes the Euclidean distance. $\mu_{\text{real}}$ and $\mu_{\text{gen}}$ are the means of the feature representations of real and generated images respectively. $\Sigma_{\text{real}}$ and $\Sigma_{\text{gen}}$ are the covariance matrices of the feature representations of real and generated images respectively. $Tr(\cdot)$ denotes the trace of a matrix.

\begin{table*}[t!]
\caption{{The performance of each method on PSV and Places2 image inpainting datasets using 40-50\% as mask ratio. The \textbf{bold} and \underline{underline} fonts respectively represent the \textbf{first} and \underline{second} place.} }
\def\arraystretch{1.2}
\label{psvres}
\footnotesize
\ignorespaces 

\begin{tabular}{|l|c|cccc|cccc|c|}
\hline

 \textbf{Method}&\textbf{Image size} &\multicolumn{4}{c|}{\textbf{Paris Street View }} & \multicolumn{4}{c|}{\textbf{Places2}}&\textbf{Para (M)}\\

\cline{3-10}
 & &\textbf{PSNR$\uparrow$} & \textbf{SSIM$\uparrow$} & \textbf{LPIPS$\downarrow$} & \textbf{FID$\downarrow$}&  \textbf{PSNR$\uparrow$} & \textbf{SSIM$\uparrow$} & \textbf{LPIPS$\downarrow$} & \textbf{FID$\downarrow$}&\\

\hline

CTN \cite{21-1}&$256\times256$&24.91&.812&-&-&  21.52&.78& -&- &21\\ \hline
ICT \cite{21-3}&$256\times256$ & & &-&-&    22.635 &0.739 &- &34.206&122 \\ \hline
MAT \cite{21-4}& $512\times512$ &- &- &-&-&  24.169 & 0.900& -&22.90&60 \\ \hline
ZITS \cite{21-5}& $256\times256$& &- &- &-&24.42 &0.870  &0.133&26.08&68\\ \hline
ZITS++ \cite{23-7}&$256\times256$& &- &- &-&24.50 &0.885  &\underline{0.118}&25.64&83\\ \hline
BAT-Fill \cite{21-6}& $256\times256$&21.76 &\underline{0.865} &0.14&63.81&  21.74 & 0.70& -&32.55&- \\ \hline
APT \cite{22-1}&$256\times256$ &24.44 &0.858 &-&-&   21.72& 0.820& -&-&- \\ \hline
T-former \cite{22-2}&$256\times256$ &25.37 &0.825 &-&46.60&  21.52 &0.770 & -&34.52& 14\\ \hline
PUT \cite{22-4}& $256\times256$& & &-&-& 22.94  &0.75 & -&31.48&- \\ \hline
TransCNN-HAE \cite{22-6}& $256\times256$ & 24.482&0.841 &-&68.791&  24.471 &0.843 & -&28.226&19 \\ \hline
Campana et al. \cite{23-1}& $256\times256$&\underline{25.821} &0.820 &\underline{0.112}&47.26&  22.317 &0.775 & 0.140&\textbf{4.640}& 17\\ \hline
SPN \cite{23-2}&$256\times256$&24.64 & 0.795&-&70.76& 22.18 & 0.763& -&\underline{4.73}&50 \\ \hline
U2AFN \cite{23-4}&$256\times256$ &-&-&-&-& 20.36  & 0.615& -&-&- \\ \hline
SWMH \cite{23-6}&$256\times256$&24.127 &0.804 &0.121&-&  - &- & -&-& \underline{6}\\ \hline
CBNet \cite{23-8}&$256\times256$& 24.72& 0.796&-&\underline{11.90}&  21.48 & 0.753& -&10.74&21 \\ \hline
CoordFill \cite{23-11}& $512\times512$&-&-&-&-&  \underline{26.365} &\underline{0.912} &\textbf{0.068}&-& \\ \hline
CMT \cite{23-16}& $256\times256$& 21.70 &0.8408& -&19.36& -&-& &-&-\\ \hline
NDMAL \cite{23-19}&$256\times256$ &- & - &-& -&  21.89 & 0.776&0.312&37.88&\textbf{4}\\ \hline
TransInpaint \cite{23-17}&$256\times256$&- & \underline{0.884}&\textbf{0.104}&\textbf{8.05}& -&- & -&-& -\\ \hline
Blind-Omni-Wav-Net \cite{23-20}&$256\times256$& \textbf{27.81} &\textbf{0.905} &-& 40.646&  \textbf{27.55} &\textbf{0.918} &-&17.521&16\\ \hline\hline
\multicolumn{11}{|c|}{\textbf{GAN}}
\\ \hline\hline
Wang et al. \cite{22-6}&$256\times256$ &\underline{25.827} &\underline{0.862} &-&114.51&  22.850 &0.838 & -&9.20&-\\ \hline
Swin-GAN \cite{23-5}& -&22.301 & 0.777&-&-&- &- & -&-& -\\ \hline
UFFC \cite{23-14}&$256\times256$&- &- &-&-& 26.41  & 0.81& 20.24&-&-\\ \hline

\end{tabular}
\end{table*}


\subsection{Results discussion}
In this section, we performed a comparison of the proposed methods in terms of the obtained results using the evaluation metrics on various image and video inpainting datasets.
For image inpainting, the most commonly used datasets were PSV, Places2, CelebA-HQ, and FHHQ. For video inpainting, the proposed method mostly commonly used the YouTube-HQ and DAVIS datasets. A comparison of the number of parameters of each model is performed. This allows the researchers assess the lightweight models regarding the computational resource challenges, especially for image and video inpainting with high-resolution images/videos.

\textbf{Evaluation on PSV and Places2 dataset}

Table \ref{psvres} shows the results obtained from transformer-based image inpainting methods with the PSNR, SSIM, LPIPS, and FID metrics applied to the PSV and Places2 datasets. In this comparison, we present the results for the mask ratio 40–50\%, which is the most used ratio for the majority of the papers. In addition, the Table shows the image input size used in the experiments.

For the PSV dataset, the obtained results show that the Blind-Omni-Wav-Net outperforms the other methods, achieving the highest PSNR values, which demonstrate it efficiency in reconstructing high-fidelity images. The methods proposed by Compana et al. and Wang et al. obtained the second best results in terms of the PSNR with a difference of two points from the Blind-Omni-Wav-Net. The majority of remaining methods exceed 24. In terms of SSIM metrics, the Blind-Omni-Wav-Net method also had the highest result, which was 0.905 better than the TransInpaint and BAT-Fill methods. This indicates that the Blind-Omni-Wav-Net preserves the structural and textural integrity of the inpainted images. On the other hand, the TransInpaint and Campana et al. methods achieved the lowest LPIPS values, reflecting superior texture and detail accuracy. In terms of the LPIPS metric, the values are close for all methods. For the FID metric, TransInpaint obtained the minimal FID score, indicating its effectiveness in generating images close to real images. This due the results obtained for SSIM, LPIPS and FID.

In the same context, the comparison was performed on the Places2 dataset using the same evaluation metrics. Almost all methods used this dataset for their experiments. The Blind-Omni-Wav-Net method obtained the highest performance in terms of the PSNR and SSIM metrics, which demonstrate the effectiveness of this method against the other proposed methods. In the second place, CoordFill reached 26.365 for PSNR and 0.912 for SSIM. This is proven by the use of attentional FFC, in addition to analyzing just the missing regions during the network process while the other regions are not analyzed and keep the same pixels values. For that, CoordFill methods can work on high- resolution images. For the other methods, including AMT ZITS, ZITS++, and TransCNN-HAE, the PSNR values were close. The CoordFill and ZITS++ methods obtained the lowest LPIPS scores, highlighting their proficiency in capturing and reproducing the complicated textures and details to different scenes in the Places2 dataset. In terms of the FID metric, the Compana et al. and SPN reached the lowest values. The differences between the obtained metrics values for each method were compared, and we found that some methods are good using one metric yet were not the best for others. This can be explained by the effectiveness of each method in specific tasks, such as preserving high-resolution quality, preserving the semantic similarity, or generating effective texture.

\begin{figure*}[t!]
\centering
  \begin{tabular}[b]{c}
    \includegraphics[width=.3\linewidth]{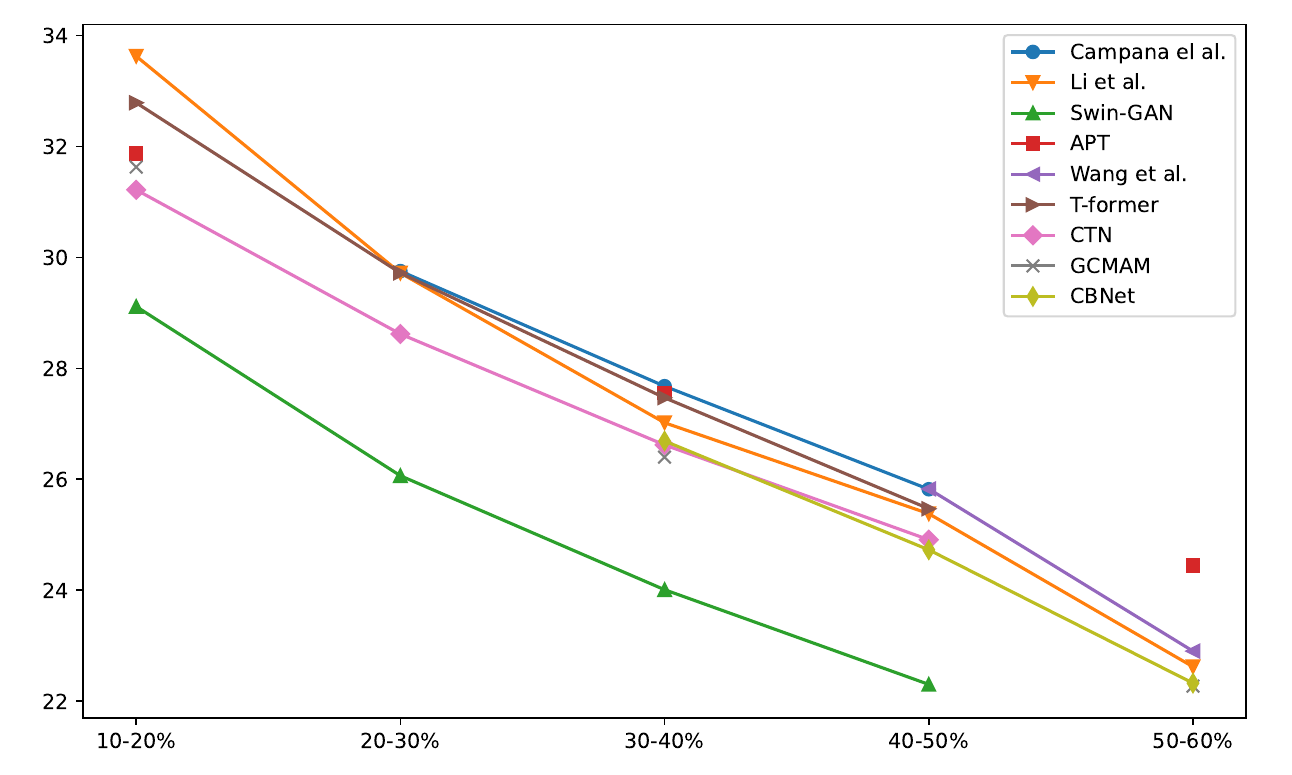}\\
    (a) PSV
  \end{tabular}
    \begin{tabular}[b]{c}
    \includegraphics[width=.3\linewidth]{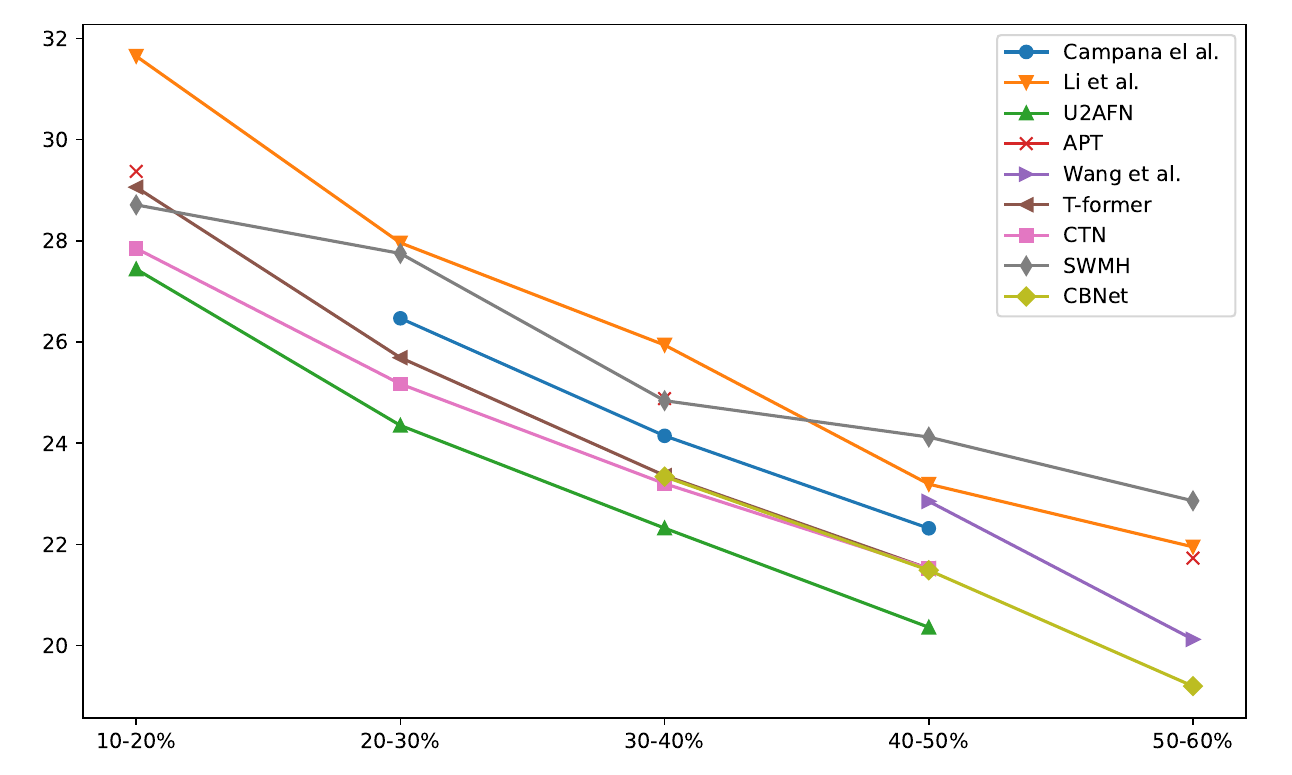}\\
    (b) Places2
  \end{tabular}
    \begin{tabular}[b]{c}
    \includegraphics[width=.3\linewidth]{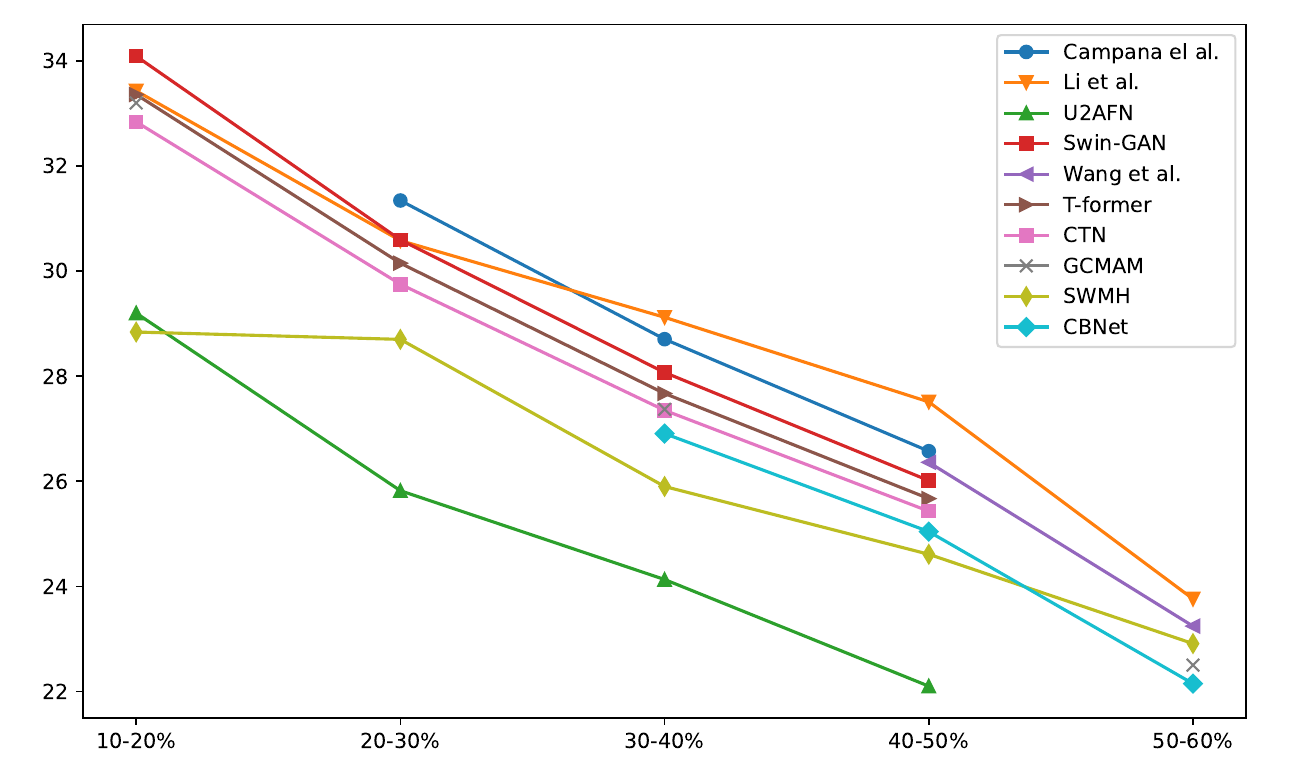}\\
    (c) CelebA
  \end{tabular}
  \caption{{Performance of image inpainting method based on mask ratio.}}
\label{mask}
\end{figure*}

The presented results in  \ref{psvres} are for the mask ratio of 40–50\%; however, some methods are performed with different mask ratios. These methods were collected and compared in terms of PSNR based on each mask ratio and illustrated in Figure \ref{mask}. We observed that the PSNR of these methods decreases when we increase the ratio of the mask. On the PSV dataset, the method by Li et al. performed well when the ratio was at 10–20\%. Furthermore, for the ratio of 50–60\%, the APT method was the best one in terms of the PSNR value. With the Places2 dataset, the same observation was found using the method by Li et al., with a 10–20\% and 20–30\%, ratio, while SwMH was the best for the 40–50\% and 50–60\% ratio. Some methods performed their experiment using two or three of these ratios, such as Compana et al., APT, and GCMAM.

\begin{table*}[t!]
\caption{{The performance of each method on the CelebA-HQ and FHHQ image inpainting datasets using 40-50\% as mask ratio. The \textbf{bold} and \underline{underline} fonts respectively represent the \textbf{first} and \underline{second} place } }
\def\arraystretch{1.2}
\label{celebares}
\footnotesize
\ignorespaces 
\centering 
\begin{tabular}{|l|c|cccc|cccc|c|}
\hline

 \textbf{Method}&\textbf{Image size}& \multicolumn{4}{c|}{\textbf{CelebA-HQ }} & \multicolumn{4}{c|}{\textbf{FHHQ}}&\textbf{Para. (M)}\\

\cline{3-10}
 & &\textbf{PSNR$\uparrow$} & \textbf{SSIM$\uparrow$} & \textbf{LPIPS$\downarrow$} & \textbf{FID$\downarrow$}&  \textbf{PSNR$\uparrow$} & \textbf{SSIM$\uparrow$} & \textbf{LPIPS$\downarrow$} & \textbf{FID$\downarrow$}&\\

\hline

CTN \cite{21-1}& $256\times256$&25.43&.909&-&-    &  -&-& -&- &21\\ \hline 
MAT \cite{21-4}& $512\times512$ &25.167 &0.917 &-&4.86&   & & -&-&60 \\ \hline
BAT-Fill \cite{21-6}& $256\times256$ &22.53& 0.843&-&11.07 &- & -& -&-&-\\ \hline
T-former \cite{22-2}& $256\times256$ & 25.67& 0.915&-&5.42&   & & -&-&14 \\ \hline
PUT \cite{22-4}& $256\times256$ & & &-&-&  24.12 &0.888 & -&19.93&-\\ \hline
TransCNN-HAE \cite{22-6}& $256\times256$ & 24.148&0.902 &-&8.783&  23.292 &0.885 & -&9.740&19\\ \hline
Campana et al. \cite{23-1}& $256\times256$ &26.572 & 0.872&\textbf{0.068}&\underline{2.253}& -&-& -&-&17\\ \hline
SPN \cite{23-2}& $256\times256$ &26.54 &0.871 &-&12.38&- & -&-&-&50\\ \hline
U2AFN \cite{23-4}& $256\times256$& 22.10&0.905 &-&-& -  & -& -&-&-\\ \hline
SWMH \cite{23-6}& $256\times256$ &24.616 &0.822 &0.103&-&  - &- & -&-&\underline{6} \\ \hline
ZITS++ \cite{23-7}& $256\times256$& &- &- &-&\underline{27.56} &\underline{0.918}  &\textbf{0.069}&\textbf{5.50}&83\\ \hline
CBNet \cite{23-8}& $256\times256$&25.04 &0.827 &-&6.307&  - & -& -&-&21 \\ \hline
CoordFill \cite{23-11}& $512\times512$ &\underline{28.756} &0.934& \underline{0.065}&-&  -&- &-&-&- \\ \hline
CMT \cite{23-16}& $256\times256$ & 23.78&0.899  &-& 5.23& -  &- & -&-&-\\ \hline
TransInpaint \cite{23-17}& $256\times256$&- & \underline{0.941}&0.079&4.46&- & -& -&-&- \\ \hline
NDMAL \cite{23-19}& $256\times256$ &23.14 & 0.858 &0.1479& 12.897&   & & -&-&\textbf{4}\\ \hline
Blind-Omni-Wav-Net \cite{23-20}& $256\times256$& \underline{28.21} &\textbf{0.951} &-&7.235&  \textbf{28.19} &\underline{0.952} & -&8.639&16\\ 
 \hline\hline

\multicolumn{11}{|c|}{\textbf{GAN}}
\\ \hline\hline
Wang et al. \cite{22-6}& $256\times256$ &26.35 &0.926 &-&11.01&  25.388 &0.914 & -&\underline{6.315}&-\\ \hline
Li et al.  \cite{23-3}& $256\times256$ &27.51 &0.89 &0.09&31.9& -  &- & -&-&-\\ \hline
Swin-GAN \cite{23-5}& $256\times256$ &26.012 & 0,870&-&-& -  & -& -&-&\\ \hline

GCAM \cite{23-18} &$512\times512$&  26.63&0.857 &-& 4.203&   & & -&-&7 \\ \hline
\end{tabular}
\end{table*}

\textbf{Evaluation on CelebA and FHHQ dataset}

Related to the PSV and Places2, the two other datasets used in different image inpainting methods using transformers which include the CelebA-HQ and FHHQ datasets, which are two datasets are of human faces. To compare the proposed methods on these two datasets, we present the obtained results using various metrics in Table \ref{celebares}. Most of the proposed methods reached convincing results in terms of the quality of the generated images and the precision of the filled region represented by SSIM metrics. For example, using the PSNR metric on the CelebA-HQ dataset, we found that 15 of the 20 methods reached a PSNR value >24, while all SSIM values were >80\%. The Blind-Omni-Wav-Net reached the best PSNR result of 28.21, followed by CoordFill. While using SSIM metric, Blind-Omni-Wav-Net and TransInpaint generated the best results. For the LPIPS metric, the Compana et al. and CoordFill methods were the best. Each one these methods work to solve a specific challenge; thus, each is best in some metrics. For example, CoordFill used a technique that can preserve the region pixels that are not missing, making it better in terms of PSNR and SSIM metrics.

Using the different mask ratios represented in Figure \ref{mask}, we can detect the differences between the methods in terms of the PSNR values. Some methods are not represented in this figure because they did not use different mask ratio, such as Blind-Omni-Wav-Net and CoorFill. From the methods that use various mask ratios, the method of Li et al. was the best in almost all ratios, except for 10–20\%. Furthermore, the SwMH method was not the best for the ratios <50\%, but the results were close to the best for the mask ratio of 50–60\%.

\begin{table*}[t!]
\caption{{The performance of each method on the YouTube-VOS and DAVIS image datasets. The \textbf{bold} and \underline{underline} fonts respectively represent the \textbf{first} and \underline{second} place.} }
\def\arraystretch{1.2}
\label{vid}
\footnotesize
\ignorespaces 
\centering 
\begin{tabular}{|l|ccccc|ccccc|}
\hline

\textbf{Method}& \multicolumn{5}{c|}{\textbf{YouTube-VOS }} & \multicolumn{5}{c|}{\textbf{DAVIS}}\\

\cline{2-5} \cline{5-11}
  &\textbf{PSNR$\uparrow$} & \textbf{SSIM$\uparrow$}  & \textbf{VFID$\downarrow$}&\textbf{$E_{warp}$$\downarrow$}   &\textbf{LPIPS$\downarrow$} & \textbf{PSNR$\uparrow$} & \textbf{SSIM$\uparrow$} & \textbf{VFID$\downarrow$} &\textbf{$E_{warp}$$\downarrow$}& \textbf{LPIPS$\downarrow$}\\

\hline

FuseFormer \cite{v21-1} &33.16 &0.9673 &0.051 &0.0875&-& 32.54 &0.9700& 0.138 & 0.1336&-\\\hline
DSTT \cite{v21-2}&32.66& 0.9646& 0.052& 0.1430& &31.75& 0.9650& 0.148 &0.1716&- \\ \hline
E2FGVI \cite{v22-1}&33.71& 0.9700& 0.046& 0.0864&& 33.01& 0.9721& 0.116& \underline{0.1315}&- \\\hline
FGT \cite{v22-2}&\underline{34.53}& \textbf{0.976}&-&-&-&33.41 &0.974 &-&-&0.023 \\\hline
DeViT \cite{v22-3}&33.42& 0.9732 &0.1429& \underline{0.049}&-& 32.43& 0.9721& 0.1663& 0.133&- \\\hline
DLFormer \cite{v22-4}&33.95& 0.970& 0.082&-&-& \underline{34.22}& \underline{0.977}& \textbf{0.062}&-&- \\\hline

DMT \cite{v23-1}&34.27 & 0.973 & 0.044&-&-& 33.82 & 0.976 & 0.104&-&- \\\hline
FGT++ \cite{v23-2}&\textbf{35.02} &\textbf{0.976}&-&-& 0.025& 33.18& 0.971 &-&-&0.028 \\\hline
ProPainter \cite{v23-5}&34.43 &\underline{0.9735}& \textbf{0.042}& 0.974&-& \textbf{34.47}& \textbf{0.9776}& \underline{0.098}& 1.187&-\\\hline
SAVIT \cite{v23-6}&33.97& 0.9727& \underline{0.043}& \textbf{0.0436}&-& 33.14& 0.9748& 0.107& \textbf{0.0673}&-\\\hline
FSTT \cite{v23-7}&34.33& 0.9731& 0.044&-&-& 33.77& 0.9756& 0.109&-&-\\\hline




\end{tabular}
\end{table*}

In the same context, some methods were evaluated on the FHHQ dataset resulted in lower values than the other datasets. The Blind-Omni-Wav-Net method achieved the best PSNR and SSIM values, while ZITS++ was the best using LPIPS and FID metrics. The number of parameters of the models can be significant for the robustness of a model, while its can be also a challenge in terms of computational resources. NDMAL and SWMH have the lowest number of parameters; however, in terms of the obtained results, they were less also than the others.

In conclusion, the obtained results of these inpainting methods across the datasets indicates not only the improvements made in the field, but also the impact of transformer-based techniques on the inpainting task. furthermore, the diversity of the methods enables the possibility of working on different aspects, such as the image quality, structural similarity, or computational efficiency, for the purpose of generating realistic images.

\begin{figure*}
\centering
\footnotesize
  \begin{tabular}[b]{c}
    \includegraphics[width=.2\textwidth]{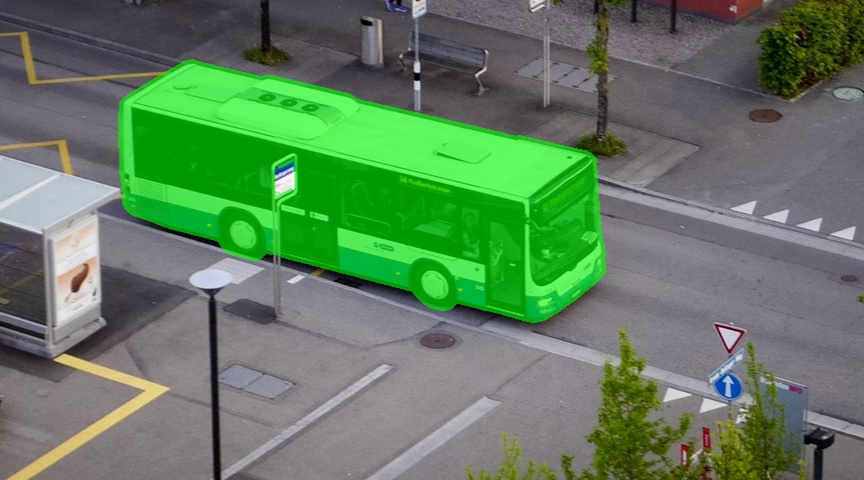}
    \includegraphics[width=.2\textwidth]{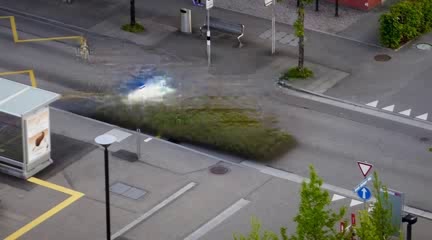}   \includegraphics[width=.2\textwidth]{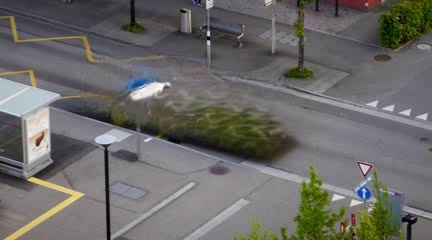}
    \includegraphics[width=.2\textwidth]{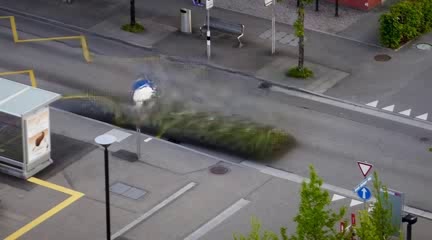}
    \includegraphics[width=.2\textwidth]{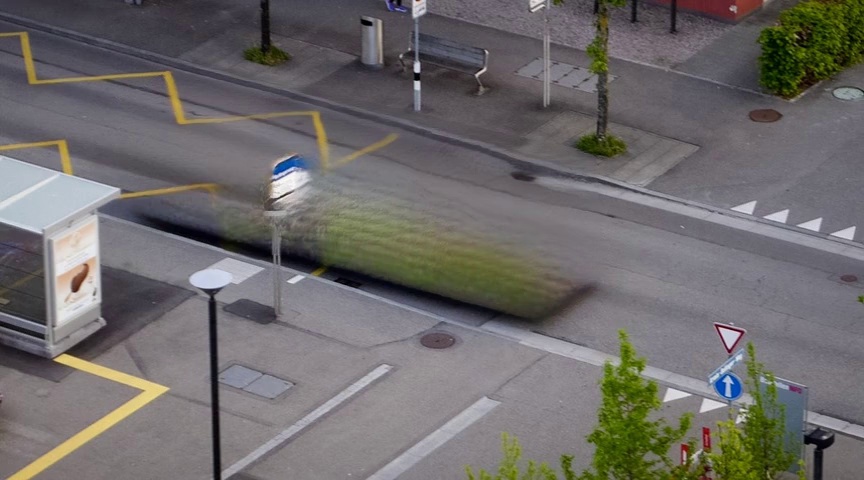}
  \end{tabular} 

  \begin{tabular}[b]{c}
    \includegraphics[width=.2\textwidth]{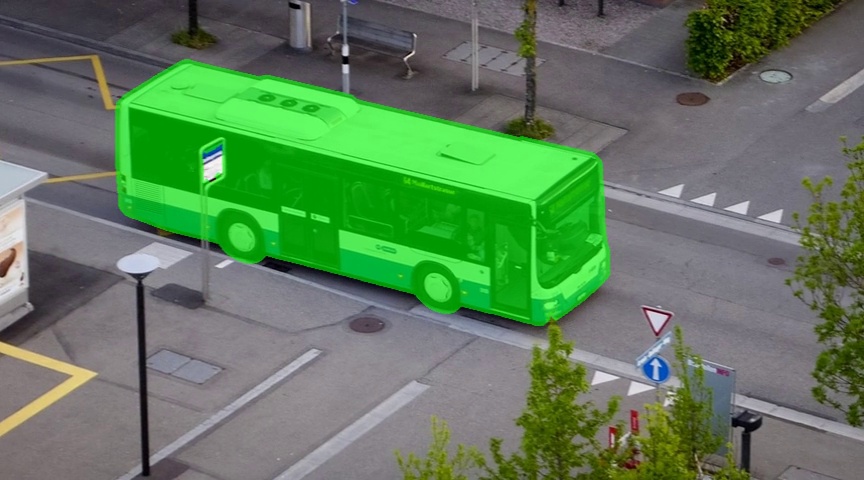}
    \includegraphics[width=.2\textwidth]{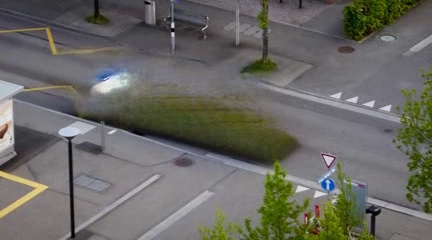}   \includegraphics[width=.2\textwidth]{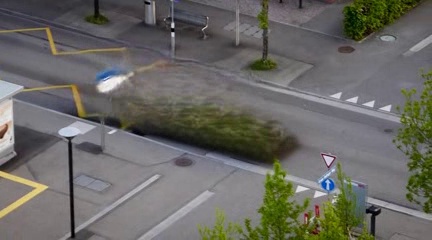}
    \includegraphics[width=.2\textwidth]{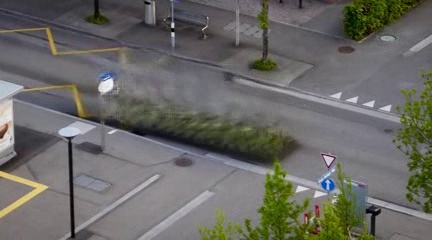}
    \includegraphics[width=.2\textwidth]{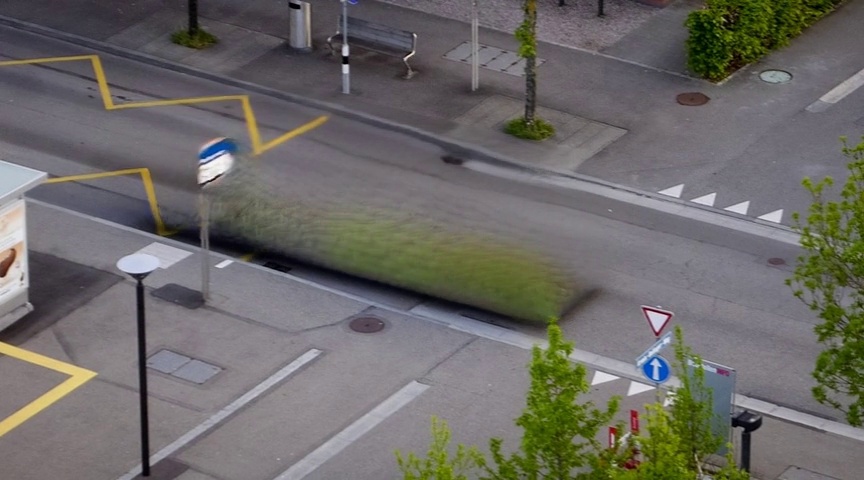}
  \end{tabular} 
  

\begin{tabular}[b]{c}
    \includegraphics[width=.2\textwidth]{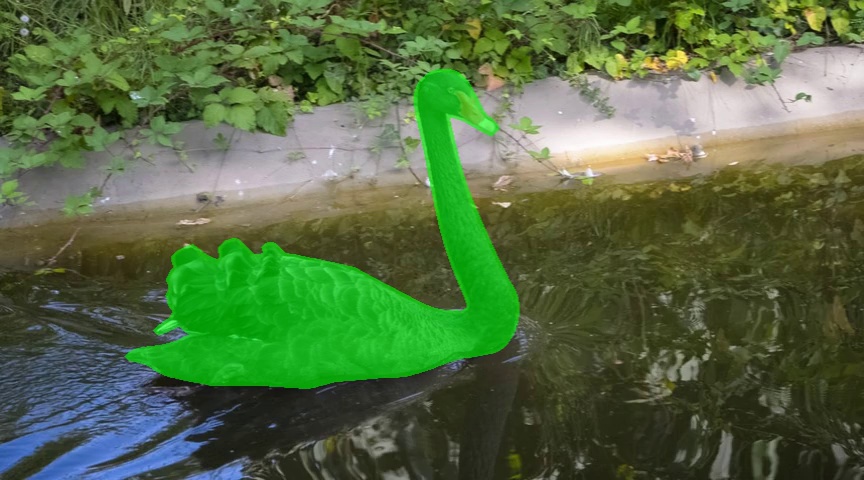}
    \includegraphics[width=.2\textwidth]{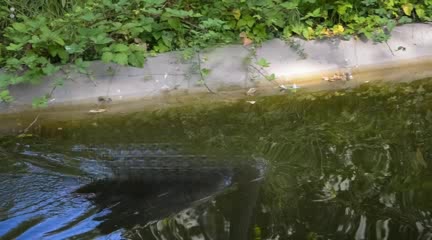}   \includegraphics[width=.2\textwidth]{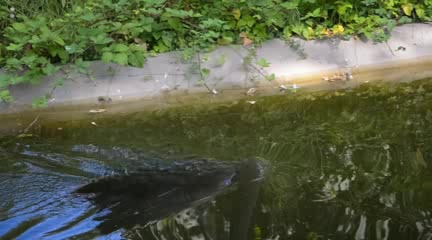}
    \includegraphics[width=.2\textwidth]{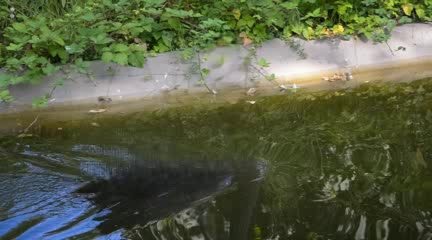}
    \includegraphics[width=.2\textwidth]{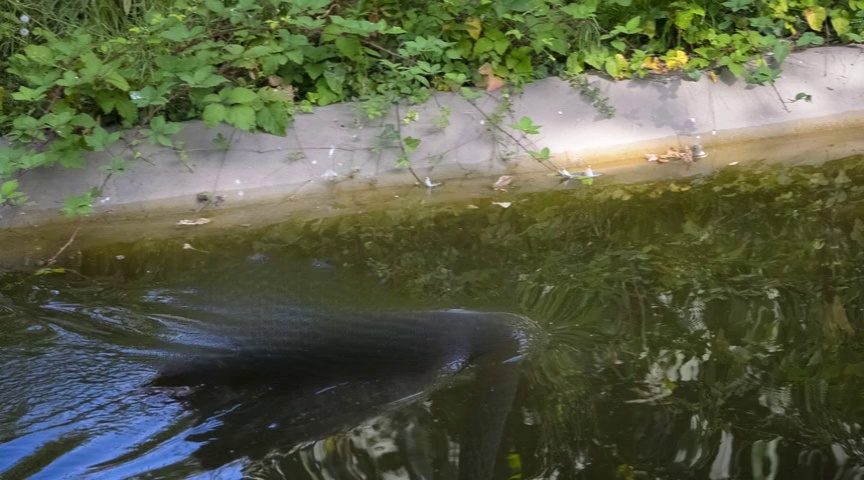}
  \end{tabular} 

\begin{tabular}[b]{c}
    \includegraphics[width=.2\textwidth]{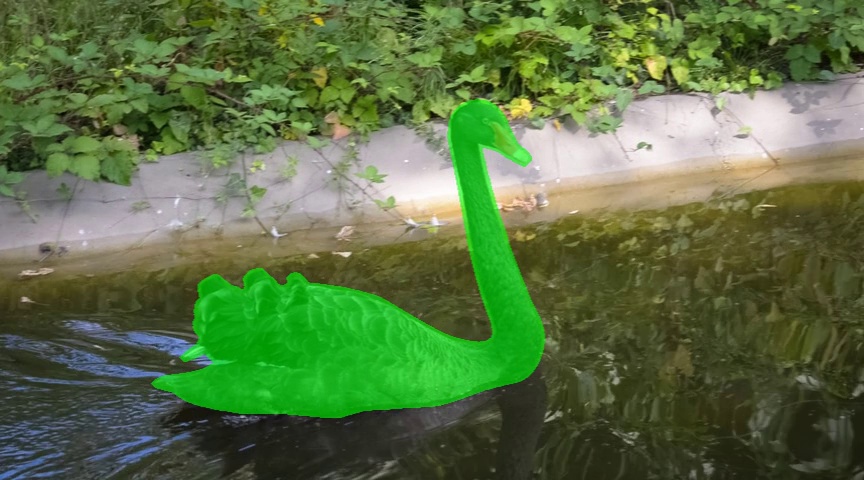}
    \includegraphics[width=.2\textwidth]{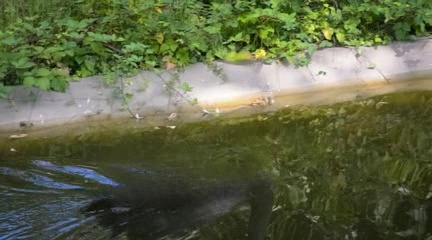}   \includegraphics[width=.2\textwidth]{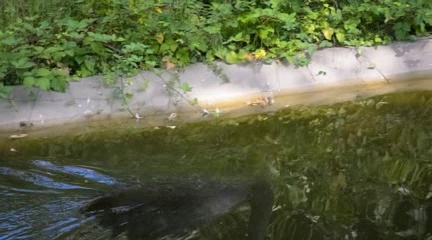}
    \includegraphics[width=.2\textwidth]{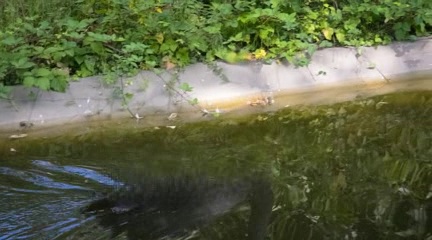}
    \includegraphics[width=.2\textwidth]{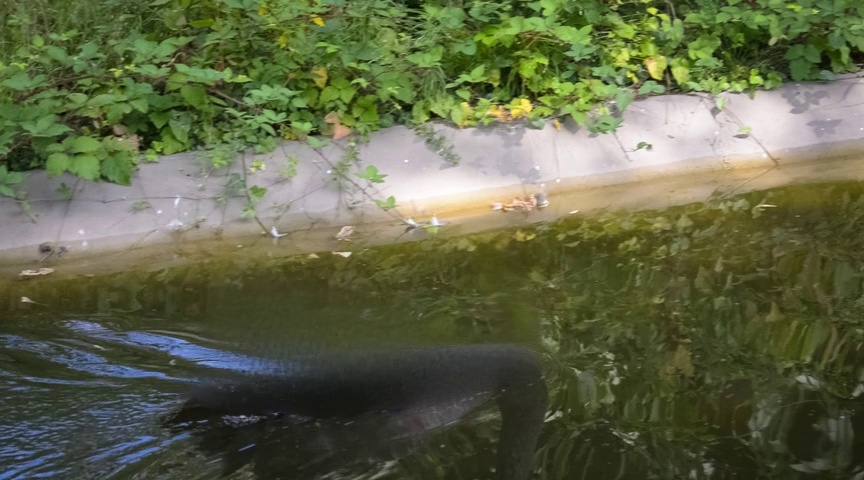}
  \end{tabular} 


\begin{tabular}[b]{c}
    \includegraphics[width=.2\textwidth]{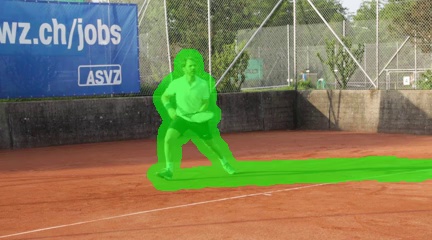}
    \includegraphics[width=.2\textwidth]{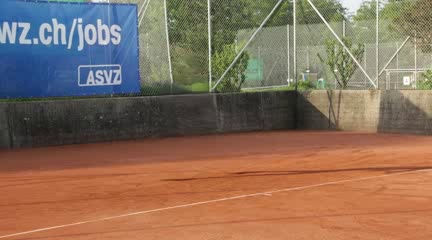}   \includegraphics[width=.2\textwidth]{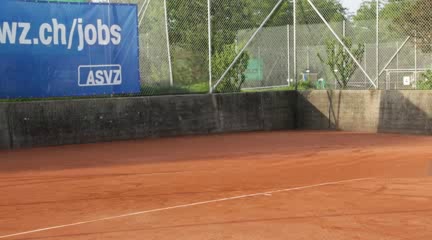}
    \includegraphics[width=.2\textwidth]{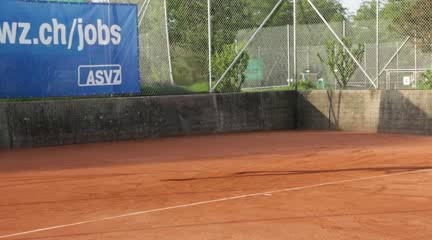}
    \includegraphics[width=.2\textwidth]{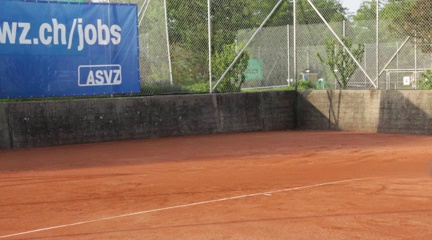}
  \end{tabular} 

\begin{tabular}[b]{c}
    \includegraphics[width=.2\textwidth]{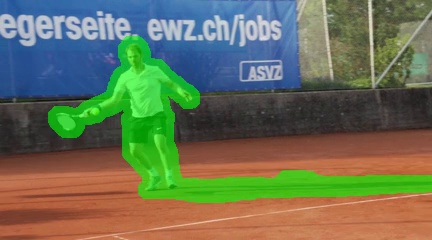}
    \includegraphics[width=.2\textwidth]{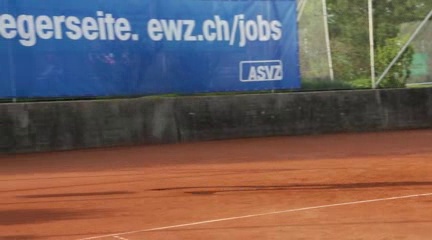}   \includegraphics[width=.2\textwidth]{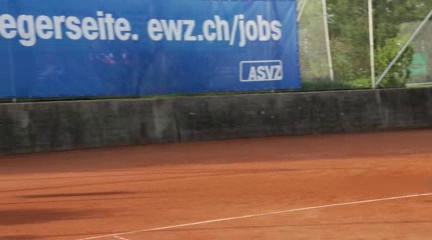}
    \includegraphics[width=.2\textwidth]{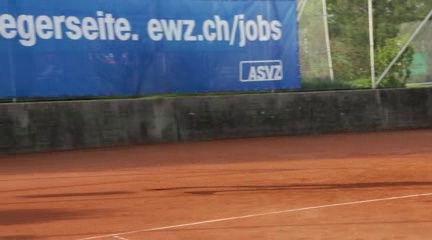}
    \includegraphics[width=.2\textwidth]{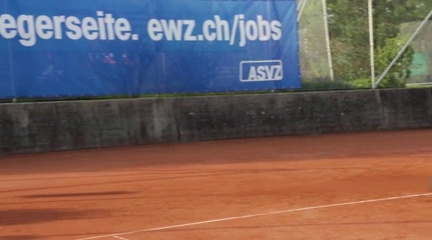}
  \end{tabular} 
\caption{The obtained results on Three videos from DAVIS dataset. First column: video fram with mask image. Second column:
DSTT. Third column: FuseFormer. Fourth column: E2FGVI. Fifth column: ProPainter.}
  \label{vidres}
\end{figure*}

\textbf{Evaluation on YouTube-VOS and DAVIS datasets}

To evaluate the proposed method for video inpainting, we performed a comparison on the most used datasets, YouTube-VOS and DAVIS. The obtained results are presented in Table \ref{vid} using different metrics, including the PSNR, SSIM, VFID, LPIPS, and $E_warp$. On both datasets, all methods used PSNR and SSIM, and VFID metrics for evaluation, while the other two other metrics used some of these methods.

For the YouTube dataset, all the obtained PSNR values exceed 33; FGT++ was the best with a value of 35.02, followed by FGT, DMT, ProPainter, and FSTT, which reached 34. The same methods reached close results in terms of the SSIM and VFID metrics. The best value for SSIM was 97\%, which reveals the improvement reached in the inpainting video with a high quality. In addition, compared to image inpainting results, the obtained results on videos are better.

Using the DAVIS dataset for video inpainting, the ProPainter and DLFormer methods reached better results in terms of the PSNR values, with a difference of 0.2 between them. The other method also achieved close PSNR values, with a difference of 1 point for most of them. The same observation was found for SSIM metrics. These results demonstrate the capability of these methods in inpainting videos, with a convincing performance. This remains true when we compare the results with the YouTube-VOS and DAVIS datasets; the results are similar even though the scenes of the two datasets are different.

To exemplify the obtained results using metrics we tested the proposed methods with source codes, including DSTT, FuseFormer, E2FGVI, and ProPainter, on three videos from the DAVIS dataset. The obtained results are illustrated in Figure \ref{vidres}. The used masks are presented in the first images and the remaining images are the obtained results. The methods do not succeed in inpainting the bus, with ProPaint inpainting it better than the other methods. For the second video the results are good. While for the third video, the method succeeds in inpainting most parts of the object; however, the shadow of the player still exists using the E2FGVI and DSTT. Overall, the transformer-based methods improve the video inpainting task in terms of quality.

\section{Image inpainting challenges}
Deep learning is a trending technology for all computer science and robotics tasks to help and assist human actions. Using artificial neural networks, which are supposed to work like a human brain, deep learning is an aspect of artificial intelligence (AI) that consists of solving the classification and recognition goals for machine learning from specific data for specific scenarios \cite{f20}. For image inpainting, the process of filling in missing or damaged areas of an image poses several challenges. These challenges can be divided into the challenges related to computer vision and architecture challenges and those related to images inpainting, such as the quality of the images. Furthermore, some challenges are related to transformer-based methods. We discuss the following set of challenges in detail.

\textbf{Preservation of Semantics:} Inpainting algorithms must preserve the semantic content of the image while filling in missing regions. The filled-in areas should blend seamlessly with the surrounding context and maintain the overall meaning of the image. Transformer-based models may struggle with preserving spatial coherence in inpainted regions, especially when dealing with complex textures or intricate structures. Ensuring smooth transitions and consistent patterns across the inpainted areas remains a challenge. In addition, inpainting requires synthesizing textures and structures to replace missing regions. Generating realistic textures that match the surrounding areas and maintaining structural coherence is essential for producing convincing inpainted results.


\textbf{Context Understanding:} Transformer models are effective at capturing long-range dependencies in sequential data, while understanding contextual information in images can be challenging \cite{21}. For image or video inpainting, understanding the global context of the image, including scene semantics and object relationships, is crucial for generating realistic and coherent inpainted results. In addition, inpainting algorithms need to accurately reconstruct missing edges and ensure smooth transitions between filled-in and original regions. This can be a challenge for deep learning models, including transformer-based models. Furthermore, there are challenges in handling the missing regions of different scales; from small scratches to large objects, removing noise or artifacts can complicate the inpainting process.

\textbf{Complexity of Architecture:} In the literature, the effective architectures used as feature extraction are generally complex, making them challenging to train, interpret, and optimize \cite{f22,f23}. Balancing model complexity with performance requirements is crucial but it can be difficult to achieve due to the other parameters, such as computational resources, especially for large scale datasets, and the number of parameters of each model (GFLOPS). In addition, training a complex architecture on some specific tasks can be more time-consuming than others. For transformer-based models based on self-attention techniques, this can make these models more complex. Thus, efficient implementation strategies and optimization techniques are required to make transformer-based inpainting methods practical for real-world use cases.

\textbf{Overfitting:} Deeper feature extraction architectures are sensitive to overfitting, where the model memorizes the training data rather than learning generalizable features \cite{f24}. Finding the best parameters, such as dropout and weight decay, can minimize the impact of this challenge; however, finding the right combination require a lot of tests and it can change from one task to another.

\textbf{Data quality requirements:} Training a CNN model requires large-scale annotated datasets, which can be expensive, time-consuming, or even unavailable for certain domains or applications. Data augmentation techniques can help to handle this challenge for some cases but may not address all scenarios for representative training data. The quality of data also represents a challenge for deep learning architecture; for example, high resolution images robustly obtain good results, but training requires a computational resource, which is another challenge. For transformer models in the case of high-resolution images, the training of the model requires dividing the image into smaller patches or applying hierarchical approaches, which can affect the quality.

\textbf{Computational Resources:} Deep-learning-based models require significant computational resources, including powerful GPUs or TPUs, for training and inference. Scaling CNNs to handle larger datasets or more complex architectures increases the demand for computational resources and limits the accessibility for researchers. The same observation was made for transformer-based models that are computationally intensive; generating real-time or interactive inpainting applications represents a challenge and requires large-scale training datasets to obtain meaningful representations.

\textbf{Domain Adaptation:} CNNs trained on specific datasets or for a task may not be suitable for different datasets or real-world environments due to domain shifts or biases. Adapting pre-trained CNNs to new domains or tasks with limited annotated data represents a challenge, especially when the target domain differs significantly from the source domain \cite{f25}. This is shown in the feature extraction models used for some specific tasks in the previous section.
For, transformer-based models, generating diverse and high-quality training data for image inpainting tasks, particularly for specific image types, can be challenging. Furthermore, it is a challenge to ensure that the model generalizes well to unseen data and various inpainting scenarios.

\section{Concluding Remarks and Future Directions}

In this paper, we reviewed several research papers on image and video inpainting techniques based on visual transformers, including their ability to capture long-range dependencies and model complex relationships within images. The proposed methods attempt to enhance the task, including efficiency and information preservation, achieving both realistic textures and structures.

Image and video inpainting have advanced significantly with the rise of deep learning, notably CNNs and GANs, which excel at filling missing or damaged regions while preserving context. Recently, transformer-based architectures have emerged as promising alternatives, leveraging self-attention mechanisms to understand global context effectively. This paper undertakes a comprehensive review, focusing on transformer-based techniques for image and video inpainting. Through a systematic categorization based on architectural configurations, types of damages, and performance metrics, we aim to demonstrate the significant progress and offer guidance to aspiring researchers in the field.

In the domain of transformer-based image and video in-
painting, a notable challenge lies in refining the model’s ability to effectively handle complex and dynamic visual contexts. This requires developing mechanisms that can seamlessly integrate temporal information in video sequences while preserving spatial coherence, thus ensuring the faithful reconstruction of missing regions. Additionally, addressing the computational cost associated with the large-scale transformer architectures demands innovative strategies for optimizing efficiency without compromising performance, thereby enabling real-time inpainting for practical applications. Furthermore, enhancing the model’s robustness to diverse and challenging inpainting scenarios, such as occlusions, irregular shapes, and varying textures, remains a critical frontier in advancing the capabilities of this transformative technology.

In terms of future research directions, several open questions remain in the realm of image and video inpainting, particularly concerning transformer-based techniques. Key avenues for further exploration include enhancing the handling of long-range dependencies to improve inpainting accuracy, investigating the performance of transformer-based approaches on diverse datasets beyond standard image and video formats to uncover new challenges and opportunities, and refining the realism and consistency of inpainted regions, especially in scenarios involving intricate textures or complex structures. Additionally, addressing temporal consistency across frames in video inpainting and ensuring robustness to various damage types, such as occlusions, corruptions, and missing data, are crucial areas for future research. Furthermore, optimizing the efficiency and scalability of transformer-based architectures for large-scale datasets or real-time applications remains an ongoing challenge. By ad- dressing these open questions, the field can advance towards more versatile, robust, and efficient inpainting solutions.

\section*{Acknowledgments}

The project is funded by the College of Information Technology (CIT), United Arab Emirates University (UAEU).

\end{document}